\documentclass[letterpaper]{article} 
\usepackage{aaai24}  
\usepackage{times}  
\usepackage{helvet}  
\usepackage{courier}  
\usepackage[hyphens]{url}  
\usepackage{graphicx} 
\usepackage{natbib}  
\usepackage{caption} 
\usepackage{algorithm}
\usepackage{algorithmic}

\usepackage{newfloat}
\usepackage{listings}
\DeclareCaptionStyle{ruled}{labelfont=normalfont,labelsep=colon,strut=off} 
\floatstyle{ruled}
\newfloat{listing}{tb}{lst}{}
\floatname{listing}{Listing}
\usepackage{xspace}
\usepackage{times}
\usepackage{epsfig}
\usepackage{graphicx}
\usepackage{amsmath}
\usepackage{amssymb}
\usepackage{booktabs}
\usepackage{algorithm}
\usepackage{algorithmic}
\usepackage{makecell}
\usepackage{animate}
\usepackage{caption}
\usepackage{tabularx}
\newcommand{\methodName}{YODA\xspace}

\title{Learn the Force We Can:\\ 
Enabling Sparse Motion Control in Multi-Object Video Generation}
\author {
    Aram Davtyan\qquad
    Paolo Favaro
}
\affiliations {
    Computer Vision Group, Institute of Informatics, University of Bern, Switzerland\\
    
    aram.davtyan@unibe.ch, paolo.favaro@unibe.ch \\
    Project website: \url{https://araachie.github.io/yoda}
}

\begin{document}

\twocolumn[{%
\renewcommand\twocolumn[1][]{#1}%
\maketitle
\setcounter{figure}{0}
\begin{center}
    \centering
    \captionsetup{type=figure}
        \begin{tabular}
            {@{}r@{\hspace{0.5mm}}c@{\hspace{0.5mm}}c@{\hspace{0.5mm}}c@{\hspace{0.5mm}}c@{\hspace{0.5mm}}c@{\hspace{0.5mm}}c@{\hspace{0.5mm}}c@{}}
            \rotatebox{90}{\makebox[2.4cm][c]{Motion \#1}} &
            {\animategraphics[width=2.4cm, autoplay, loop]{3}{Figures/bair_index_9_robot_int/image_}{0}{10}} & {\includegraphics[width=2.4cm]{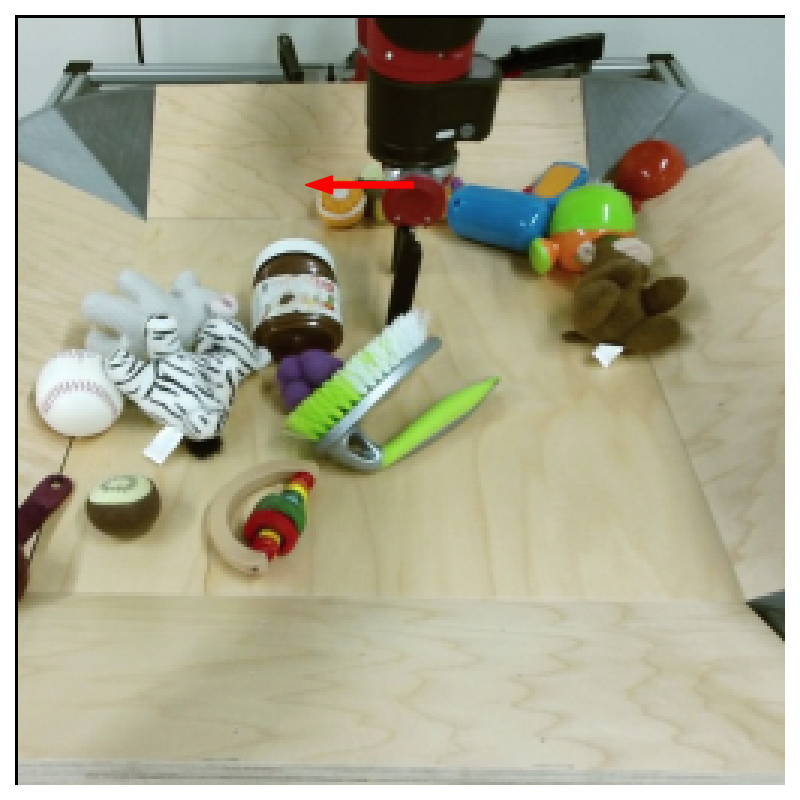}} & 
            {\includegraphics[width=2.4cm]{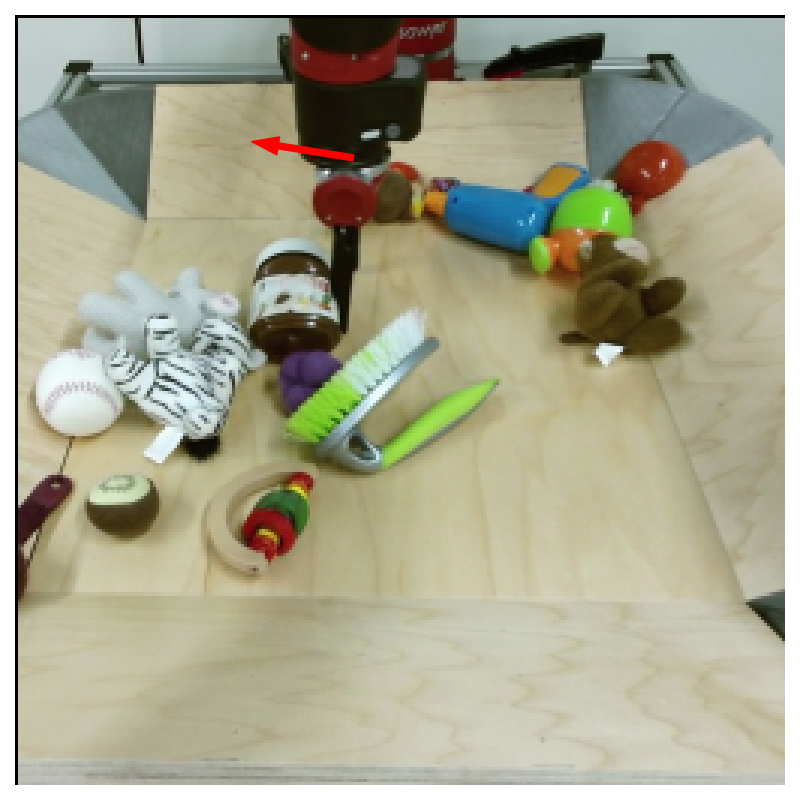}} & 
            {\includegraphics[width=2.4cm]{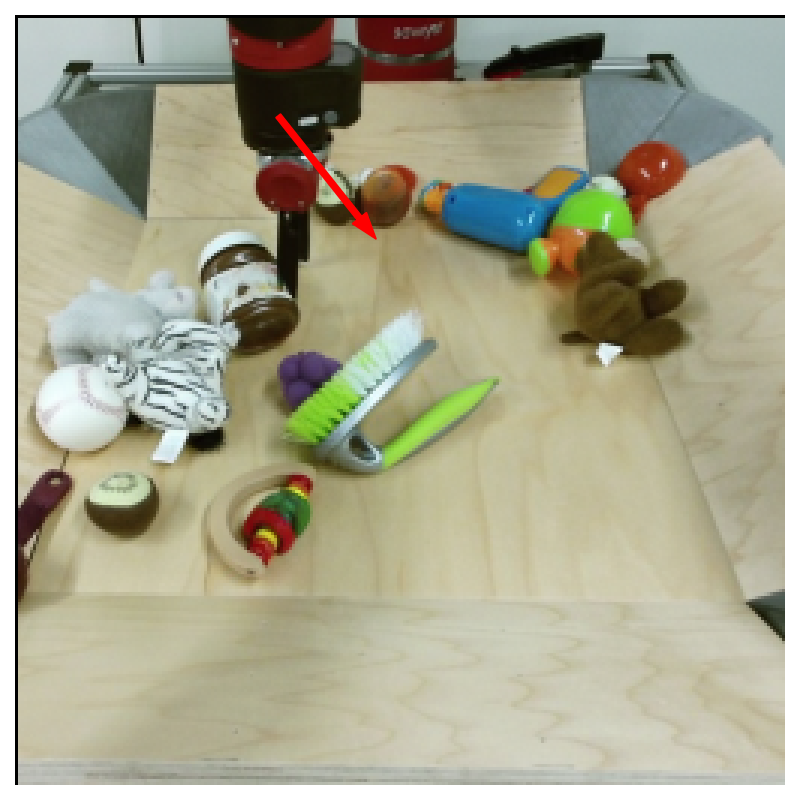}} &
            {\includegraphics[width=2.4cm]{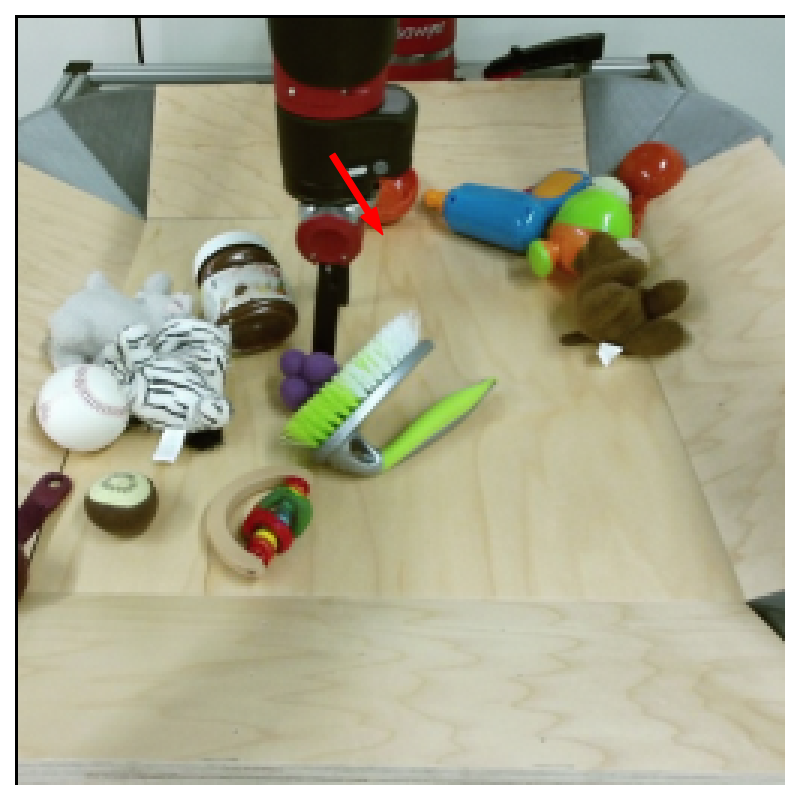}} & 
            {\includegraphics[width=2.4cm]{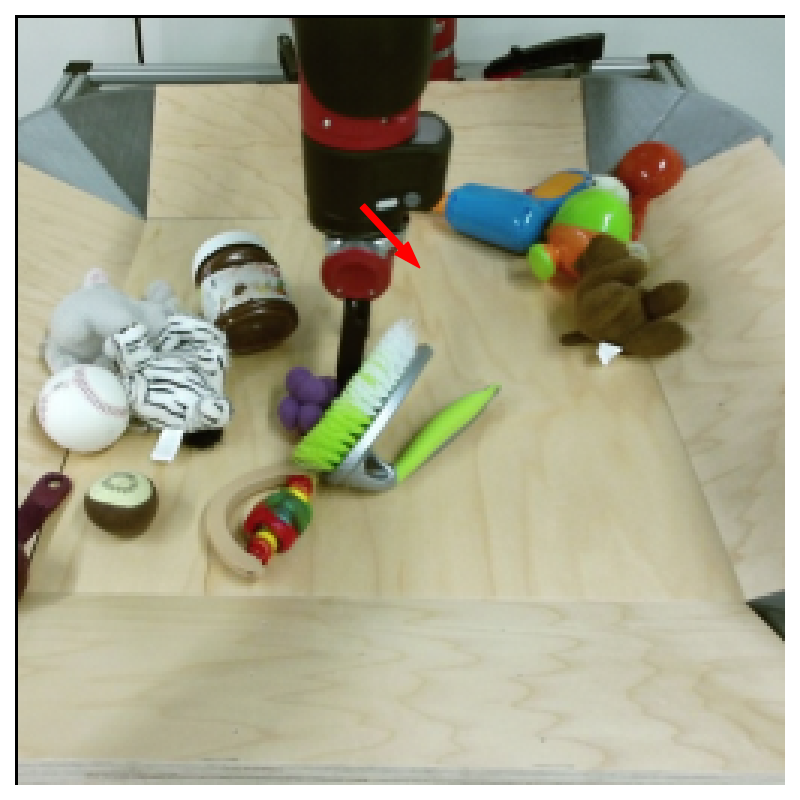}} & 
            {\includegraphics[width=2.4cm]{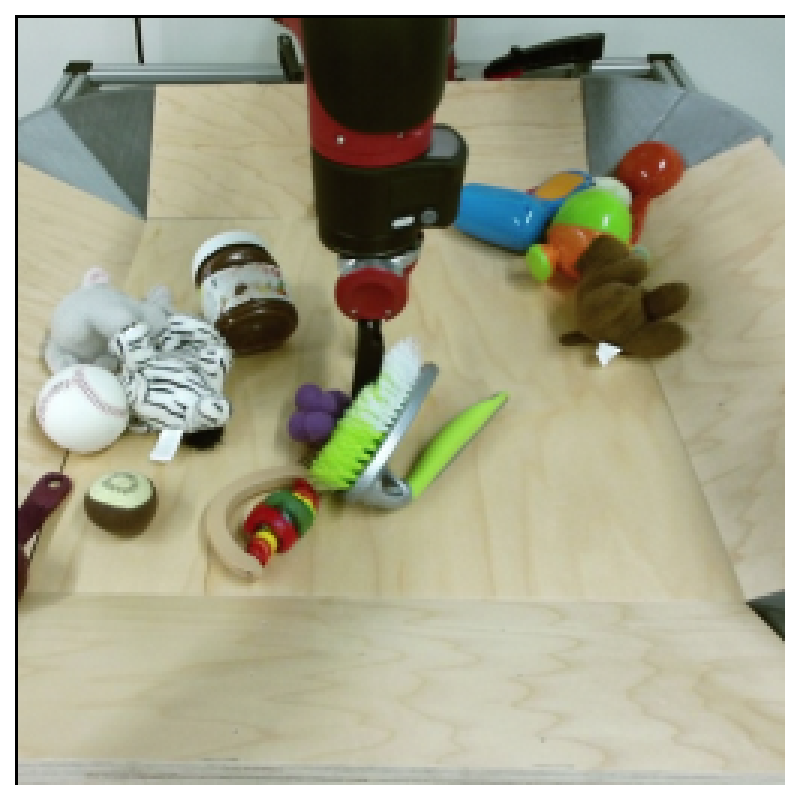}} \\
            \rotatebox{90}{\makebox[2.4cm][c]{Motion \#2}} &
            {\animategraphics[width=2.4cm, autoplay, loop]{3}{Figures/bair_rotate/0000}{0}{7}} & {\includegraphics[width=2.4cm]{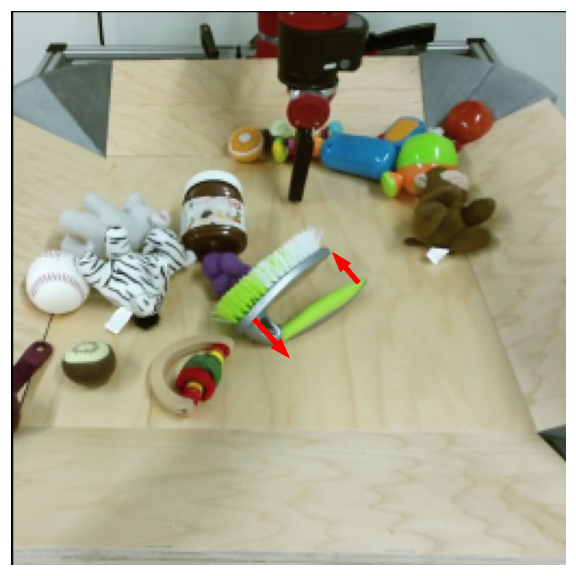}} & 
            {\includegraphics[width=2.4cm]{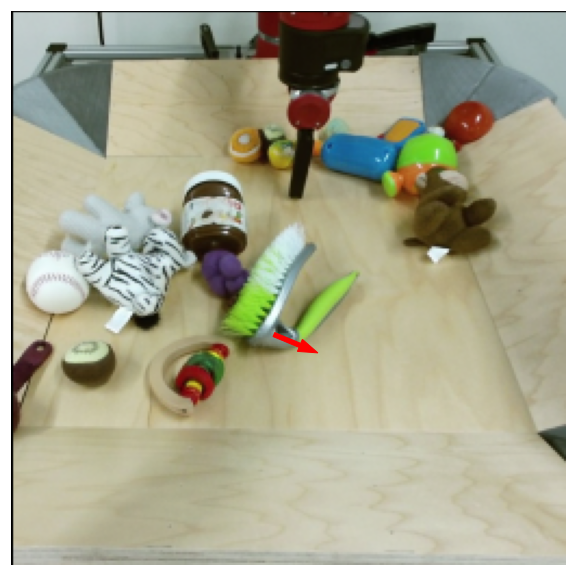}} & 
            {\includegraphics[width=2.4cm]{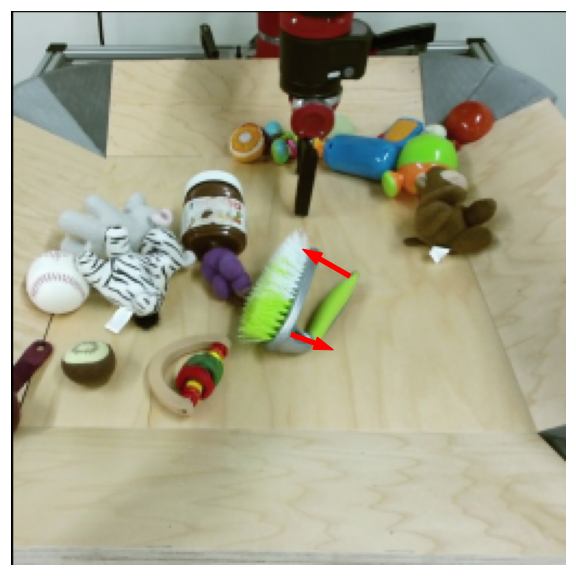}} &
            {\includegraphics[width=2.4cm]{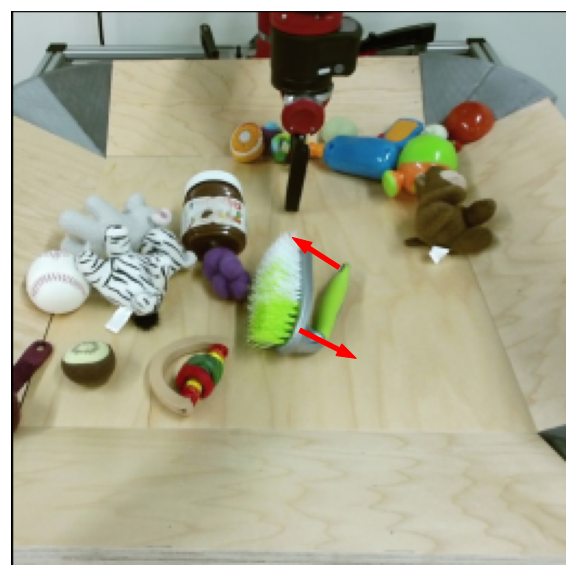}} & 
            {\includegraphics[width=2.4cm]{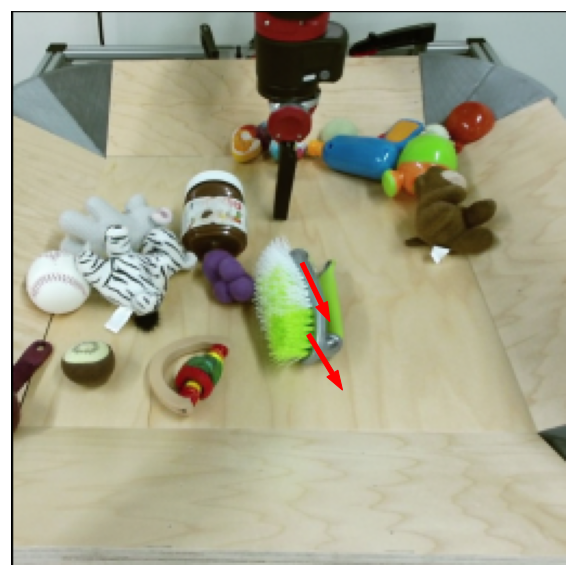}} & 
            {\includegraphics[width=2.4cm]{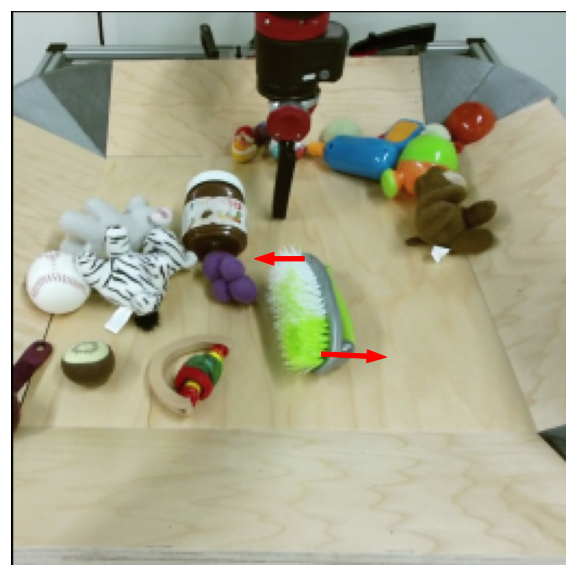}} \\
        \end{tabular}
    \captionof{figure}{
    Examples of videos generated through controlled motions by \methodName on the BAIR dataset. Both videos are generated autoregressively by starting from the same single image and then by providing control inputs in the form of 2D shifts (shown as red arrows superimposed to the frames). 
    To play the videos in the first column on the left, view the paper with Acrobat Reader. \label{fig:bair_teaser}}
\end{center}
}]
\begin{abstract}
We propose a novel unsupervised method to autoregressively generate  videos from a single frame and a sparse motion input. Our trained model can generate unseen realistic object-to-object interactions. 
Although our model has never been given the explicit segmentation and motion of each object in the scene during training, it is able to implicitly separate their dynamics and extents.
Key components in our method are the randomized conditioning scheme, the encoding of the input motion control, and the randomized and sparse sampling to enable generalization to out of distribution but realistic correlations.
Our model, which we call \methodName, has therefore the ability to move objects without physically touching them. 
Through extensive qualitative and quantitative evaluations on several datasets, we show that \methodName is on par with or better than state of the art video generation prior work in terms of both controllability and video quality. 
\end{abstract}

\section{Introduction}

Imagining the potential outcomes of own and others' actions is considered the most powerful among the three levels of cognitive abilities that an intelligent system must master according to \cite{pearl2018book}. Imagining is particularly valuable. It allows organisms to understand and analyse possible futures, which enables them to plan ahead and significantly increases their chances of survival. Thus, including the capability of imagining seems a natural choice for the design of intelligent systems. 


In this work, we aim at building models for video generation that learn to imagine the outcome of sparse control inputs in multi-object scenes. Learning is allowed only through the passive observation of a video dataset, \emph{i.e.}, without the ability to explicitly interact with the scene, and without any  manual annotation.
We aim to build a model that can show some degree of generalization from the training data, \emph{i.e.}, that can generate data that was not observed before, that is directly related to the control input, and that is plausible, as shown in Figure~\ref{fig:bair_teaser}. One possible choice of controls expressed in natural language could take the form ``Move \texttt{this} object at \texttt{this} location''. However, such controls assume that the objects in an image are already known. Since we do not use annotations, it is quite difficult to obtain an explicit scene decomposition into objects. 
Therefore, we opt for representing the controls as 2D shifts at pixels and let the model implicitly learn the extents of the objects during training by observing correlated motion of the pixels in real videos. Thus, the control inputs we consider correspond instead to sentences of this form: ``Move \texttt{this} pixel (and the corresponding underlying object) at \texttt{this} location''.
The model is fed with the current frame, \emph{context} frames (a subset of the past frames to enable memory), and a set of control inputs. The output is a generated subsequent frame that shows how the input frame would change under the specified motion.
We train our model \textbf{without} specifying what objects are (\emph{i.e.}, we do not use information about object categories), or where they are (\emph{i.e.}, we do not rely on bounding boxes, landmarks/point annotations or segmentation masks), or how they interact (\emph{i.e.}, we do not make use of information on object relationships or action categories or textural descriptions of the scenes). 
Given two subsequent frames (the current and following one) from a video in our training data, we obtain the motion control input by sampling the estimated optical flow at a few (typically $5$) locations. 
To generate frames we use flow matching \cite{lipman2022flow, liu2022flow, albergo2022building}, where the model is conditioned on the current and past frames \cite{Davtyan_2023_ICCV}, and feed the motion control input through cross-attention layers in a transformer architecture.

As shown in Figure~\ref{fig:bair_teaser}, an emerging property of our proposed approach is that the model learns the physical extent of objects in the scene without ever requiring explicit supervision for it. For example, although the control is applied to the handle of the brush in the second row, the motion is applied correctly to the whole brush, and to the whole brush only. A second learned property can be observed on the first row: When the robot arm is driven towards other objects, it interacts with them realistically. In the second row, we can observe a third remarkable capability that the model has learned. The generated video shows that we can directly rotate a single object, without using the robot arm to do so. This is a video that has never been observed in the dataset (all objects are moved directly by the arm or indirectly via other objects). Because of this ability of moving objects without touching them, we call our approach \methodName. It does demonstrate empirically that the model has the ability to \emph{imagine} novel plausible outcomes when the reality is modified in ways that were not observed before.

Our contributions can be summarized as follows
\begin{enumerate}
    \item We introduce a model for controllable video synthesis that is trained in a completely unsupervised fashion, is not domain-specific, and can scale up to large datasets;
    \item We introduce an effective way to embed motion information and to feed it to the model, and show analysis to understand the impact of sampling and the use of sparsity of the motion field. We conclude that all these components are crucial to the correct learning of object interactions and the disentanglement of object motion control; 
    \item We demonstrate for the first time multi-object interactions in the unsupervised setting on real data, which has not been shown in other state of the art methods \cite{menapace2021playable,menapace2022playable,davtyan2022glass,huang2022layered,blattmann2021understanding,blattmann2021ipoke}.
\end{enumerate}
\section{Prior Work}

\textbf{Video generation.} An increased interest in video generation has followed the success of generative models for images~\cite{karras2021alias, esser2021taming,  Dhariwal2021DiffusionMB, rombach2022high}. In contrast to image generation, video generation is plagued by problems such as rendering realistic motion, capturing diversity (i.e., modeling the stochasticity of the future outcomes) and, most importantly, managing the high computational and storage requirements. Conventional approaches to video generation are autoregressive RNN-based models~\cite{babaeizadeh2021fitvid, blattmann2021understanding}. 
Other models instead directly generate a predefined number of frames~\cite{ho2022video, blattmann2021ipoke, singer2022make}. Variability (stochasticity) of the generated sequences has been tackled with GANs~\cite{Clark2019AdversarialVG,Luc2020TransformationbasedAV}, variational approaches~\cite{Babaeizadeh2018StochasticVV, denton2018stochastic,babaeizadeh2021fitvid,Lee2021RevisitingHA}, 
Transformers~\cite{le2021ccvs,rakhimov2020latent,gupta2022maskvit, yan2021videogpt} and diffusion-based approaches~\cite{ho2022video, voleti2022mcvd,hoppe2022diffusion,harvey2022flexible}. The recently proposed autoregressive method RIVER~\cite{Davtyan_2023_ICCV} deals with the stochastic nature of the generative process through flow matching~\cite{lipman2022flow}. Among all above approaches, \methodName uses RIVER as a backbone, because of its efficiency and ease of training. 

\noindent\textbf{Controllable video generation} work mostly differs in the nature of the control signals. Control can be defined per frame \cite{chiappa2017recurrent,finn2016unsupervised,kim2020learning,nunes2020action,oh2015action, hu2021learning}, or as a global label \cite{wang2020imaginator, singer2022make}. Some of them are obtained via supervision \cite{chiappa2017recurrent,finn2016unsupervised,kim2020learning,nunes2020action,oh2015action}, or discovered in an unsupervised manner \cite{menapace2021playable, davtyan2022glass, huang2022layered, rybkin2018learning}. For instance, CADDY~\cite{menapace2021playable} learns a discrete action code of the agent that moves in the videos. Another model, GLASS~\cite{davtyan2022glass} decouples the actions into global and local ones, where global actions, as in \methodName, are represented with 2D shifts, while local actions are discrete action codes, as in \cite{menapace2021playable}. In \cite{huang2022layered} the authors explicitly separate the foreground agent from the background and condition the generation on the transformations of the segmentation mask. However, all these models are restricted to single agent videos, while \methodName successfully models multiple objects and their interactions. \cite{hu2021learning, hao2018controllable, blattmann2021understanding, blattmann2021ipoke} are the most similar works to ours as they specify motion control at the pixel-level. However, \cite{hu2021learning} leverages a pre-trained object detector to obtain the ground truth control. \cite{hao2018controllable} is based on warping and therefore does not incorporate memory to model long-range consequences of actions. II2V~\cite{blattmann2021understanding} uses a hierarchical RNN to allow modeling higher-order details, but focuses on deterministic prediction. iPOKE~\cite{blattmann2021ipoke} aims to model stochasticity via a conditional invertible neural network, but has to sacrifice the ability to generate long videos and to intervene into the generation process at any timestamp. None of those works has demonstrated controllable video generation on multi-object real scenes. 

\noindent\textbf{Multi-object scenes and interactions.} Modeling multiple objects in videos and especially their interactions is an extremely difficult task. It either requires expensive human annotations~\cite{hu2022make} or is still limited to simple synthetic scenes~\cite{wu2021generative, schmeckpeper2021object, janner2018reasoning}. \cite{menapace2022playable, yan2022patch} allow for multi-agent control, but leverage ground truth bounding boxes during the training. 
\methodName in turn is an autoregressive generative model for controllable video generation from sparse motion controls that i) efficiently takes memory into account to simulate long-range outcomes of the actions, ii) models stochasticity of the future, iii) does not require human annotation for obtaining the control signal(s), and iv) demonstrates controllability on a complex multi-object real dataset.

\section{Training \methodName}

We denote with $\mathbf{x} = \{x^1, \dots, x^N\}$ an RGB video that contains $N$ frames, where $x^i \in \mathbb{R}^{3\times H\times W}, i = 1, \dots, N$, and $H$ and $W$ are the height and the width of the frames respectively. The goal is to build a controllable video prediction method that allows us to manipulate separate objects in the scene. We formulate this goal as that of approximating a sampler from the following conditional distribution
\begin{align}
    p(x^{k + 1} | x^{k}, x^{k-1}, \dots, x^1, a^{k}) \label{eq:dec},
\end{align}
for $k<N$ and where $a^k$ denotes the motion control input. $a^k$ specifies the desired shifts at a set of pixels (including the special cases with a single pixel or none). Our ultimate objective is to ensure through training that this control implicitly defines the shift(s) for the object(s) containing the selected pixel(s). 
The conditioning in eq.~\eqref{eq:dec} allows an autoregressive generative process at inference time, where the next frame $x^{k+1}$ in a generated video is sampled conditioned on the current frame $x^k$, the previously generated frames $x^{k-1}, \dots, x^1$ and the current control $a^k$. To model the conditional distribution in eq.~\eqref{eq:dec}, we use RIVER~\cite{Davtyan_2023_ICCV}. This is a recently proposed video prediction method based on conditional flow matching~\cite{lipman2022flow}. 
We chose RIVER due to its simplicity and training efficiency compared to conventional RNNs~\cite{babaeizadeh2021fitvid, 10.1162/neco.1997.9.8.1735, Babaeizadeh2018StochasticVV, denton2018stochastic, lee2018stochastic} and Transformers~\cite{NIPS2017_3f5ee243, weissenborn2019scaling, le2021ccvs,rakhimov2020latent,gupta2022maskvit,seo2021autoregressive} for video prediction. 
For completeness, we briefly introduce Flow matching and RIVER in section~\ref{sec:river}. In section~\ref{sec:force} we show how the latter is adapted to handle control. In section~\ref{sec:control}, we focus on how the control signals are obtained and encoded. 

\subsection{Preliminaries: Flow Matching and RIVER}\label{sec:river}

\noindent\textbf{Flow matching} \cite{lipman2022flow, liu2022flow, albergo2022building} was introduced as a simpler, more general and more efficient alternative to diffusion models~\cite{Ho2020DenoisingDP}. 
The goal is to build an approximate sampler from the unknown data distribution $q(y)$, given a training set of samples of $y$. This is formalized as a continuous normalizing flow~\cite{chen2018neural} via the following ordinary differential equation 
\begin{align}\label{eq:ode}
    \dot \phi_t(y) = v_t(\phi_t(y)), \quad \phi_0(y) = y.
\end{align}
Eq.~\eqref{eq:ode} defines a flow $\phi_t(y): [0, 1] \times \mathbb{R}^d \rightarrow \mathbb{R}^d$ that pushes $p_0(y) = {\cal N}(y \,|\, 0, 1)$ towards the distribution $p_1(y) \approx q(y)$ along the vector field $v_t(y): [0, 1] \times \mathbb{R}^d \rightarrow \mathbb{R}^d$.
Remarkably, \cite{lipman2022flow} shows that one can obtain $v_t(y)$ by solving
\begin{align}\label{eq:cfm}
    \min_{v_t} \;\mathbb{E}_{t, p_t(y \,|\, y_1), q(y_1)} \| v_t(y) - u_t(y \,|\, y_1) \|^2,
\end{align}
where one can explicitly define the vector field $u_t(y | y_1)$ and its corresponding probability density path 
$p_t(y | y_1)$, with $y_1 \sim q(y)$. A particularly simple choice \cite{lipman2022flow} is the Gaussian probability path $p_t(y \,|\, y_1) = {\cal N}(y \,|\, \mu_t(y_1), \sigma^2_t(y_1))$, with $\mu_0(y_1) = 0, \mu_1(y_1) = y_1, \sigma_0(y_1) = 1, \sigma_1(y_1) = \sigma_{\text{min}}$. The corresponding target vector field is then given by
\begin{align}\label{eq:u}
    u_t(y \,|\, y_1) = \frac{y_1 - (1 - \sigma_{\text{min}}) y}{1 - (1 - \sigma_{\text{min}}) t}.
\end{align}
Sampling from the model that was trained to optimize~\eqref{eq:cfm} can be obtained by first sampling $y_0 \sim {\cal N}(y \,|\, 0, 1)$ and then by numerically solving eq.~\eqref{eq:ode} to obtain $y_1 = \phi_1(y_0)$. 

\vspace{5pt}\noindent\textbf{RIVER} \cite{Davtyan_2023_ICCV} is an extension of the above procedure to the video prediction task with a computationally efficient conditioning scheme on past frames.
The training objective of RIVER is given by
\begin{align}
    {\cal L}_{\text{R}}(\theta) = \| v_t(x \,|\, x^{\tau-1}, x^{c}, \tau - c \,; \theta) - u_t(x \,|\, x^{\tau}) \|^2, \label{eq:river}
\end{align}
where $v_t$ is a network with parameters $\theta$, $x^{\tau}$ is a frame randomly sampled from the training video, $c$ is an index randomly sampled uniformly in the range $\{1, \dots, \tau - 2\}$ and $u_t$ is calculated with eq.~\eqref{eq:u}. An additional information provided to $v_t$ is the time interval $\tau-c$ between the target frame $x^\tau$ and the past frame $x^c$, which we call the \emph{context frame}.

At test time, during the integration of eq.~\eqref{eq:ode}, a new context frame $x^c$ is sampled at each step $t$. This procedure enables to condition the generation of the next frame on the whole past. To further speed up the training and enable high-resolution video synthesis, RIVER works in the latent space of a pretrained VQGAN~\cite{esser2021taming}. That is, instead of $x^\tau, x^{\tau - 1}, x^c$ in eq.~\eqref{eq:river} one should write $z^\tau, z^{\tau - 1}, z^c$, where $z^i$ is the VQ latent code of the $i$-th frame \cite{Davtyan_2023_ICCV}. Since the use of VQGAN encoding is an optional and separate procedure, we simply use $x$ in our notation.

\subsection{Learning to Master the Force}\label{sec:force}


We now show how to incorporate control into eq.~\eqref{eq:river} to build a sampler for the conditioning probability in eq.~\eqref{eq:dec}. To do so, we make $v_t$ depend on $a^{\tau - 1}$, which is the motion control at time $\tau - 1$, as shown in the following objective
\begin{align}
    {\cal L}_{\text{F}}(\theta) = \| v_t(x \,|\, x^{\tau-1}, x^{c}, \tau - c, a^{\tau - 1} \,; \theta) - u_t(x \,|\, x^{\tau}) \|^2.\label{eq:force}
\end{align}
In practice, we implement this conditioning by substituting the bottleneck in the self-attention layers of the U-ViT~\cite{bao2022all} architecture of RIVER with cross-attention blocks (for detailed description of the architecture, see supplementary material). The control inputs $a^{\tau - 1}$ are obtained by splitting the image domain into a grid of tiles, so that a motion control can be specified in each tile via a code, and then be fed as keys and values to the cross-attention layer. More details on $a^{\tau-1}$ are provided in the next section.

Inspired by the classifier-free guidance for diffusion models~\cite{ho2022classifier}, we propose to switch off the conditioning on both the context and motion control at every iteration, with some probability $\pi$ (see Algorithm~\ref{alg:force}). This is done by using noise as the code corresponding to a switched off context or a motion control. A typical value for $\pi$ in our experiments is 0.5. This procedure yields two important outcomes: 1) a  model that makes use of the conditioning signals and 2) a model that can generate a video by starting from a single frame. As evidence of the outcome 1), note that the conditioning on both the context frame $x^c$ and the control $a^{\tau - 1}$ is often redundant, because in many instances the future frame can be reliably predicted given only one of the two conditioning signals. In these cases, the model could learn to use only one of them. By randomly switching off the conditions, we force the model to always use both of them, as it does not know which one might be missing. 
In addition, this training procedure enables the model to change control on the fly.
To see 2), consider that when the model generates the first predicted frame, there is no valid context frame and our procedure allows us to replace the context frame with noise. In alternative, one would have to duplicate the first frame, for example, but this would result in an undesirable sampling bias.
\begin{algorithm}[t]
    \caption{Training of \methodName}\label{alg:force}
    \begin{algorithmic}[1]
        \STATE Input: dataset of videos $D$, number of iterations $N$;
        \FOR{$i$ in range(1, $N$)}
        \STATE Sample a video from the dataset $\mathbf{x} \sim D$ of length $L$;
        \STATE Choose a random target frame $x^{\tau}, \tau \in \{3, \dots L\}$
        \STATE Sample a timestamp $t \sim U[0, 1]$;
        \STATE Sample a noisy observation $x \sim p_t(x \,|\, x^{\tau})$;
        \STATE Calculate $u_t(x \,|\, x^{\tau})$ according to eq.~\eqref{eq:u};
        \STATE Sample a condition frame $x^{c}, c \in \{1, \dots \tau - 2\}$;
        \STATE Compute the control $a^{\tau - 1}$ (see sec.~\ref{sec:control});
        \STATE With probability $\pi$ set $x^c$ to noise;
        \STATE With probability $\pi$ set $a^{\tau - 1}$ to noise;
        \STATE Take a gradient descent update $\theta \leftarrow \theta - \alpha\nabla_{\theta} \mathcal{L}_\text{F}(\theta)$
        with learning rate $\alpha>0$;
        \ENDFOR
    \end{algorithmic}
\end{algorithm}

\subsection{Force Embeddings}\label{sec:control}

Ideally, $a^{\tau-1}$ could encode detailed motion information for the objects in the scene. For instance, $a^{\tau-1}$ could describe that an object is rotating or pressing against another object or walking (in the case of a person). Such supervision could potentially provide the ability to control the video generation in detail and to generalize well to unseen object motion combinations. However, obtaining such ground truth control signals requires costly large-scale manual annotation. Similarly to \cite{blattmann2021ipoke}, we avoid such costs by leveraging optical flow. Essentially, instead of using a costly and detailed motion representation, we use a simpler one that can be computed automatically and at a large scale.

Given an optical flow $w^\tau \in \mathbb{R}^{2\times H \times W}$ between the frames $x^\tau$ and $x^{\tau + 1}$ (obtained with a pretrained RAFT~\cite{teed2020raft}), we define a probability density function $p(i, j) \propto \|w^\tau_{ij}\|^2$ over the image domain $\Omega$, with $(i,j)\in\Omega$. Then, we randomly sample a sparse set $\mathcal{S}\subset \Omega$ of $n_c=|\mathcal{S}|$ pixel locations from $p$.
This distribution makes it more likely that pixels of moving objects will be selected. However, in contrast to \cite{blattmann2021ipoke} we do not introduce additional restrictions to the sampling or explicitly define the background. This is an essential difference, because in multi-object scenes, objects that belong to the background in one video might be moving in another. Thus, in our case, one cannot use the magnitude of the optical flow to separate objects from the background in each video.

\begin{figure}[t]
    \centering
    \includegraphics[width=0.9\linewidth, trim=24cm 14cm 18.4cm 9cm, clip]{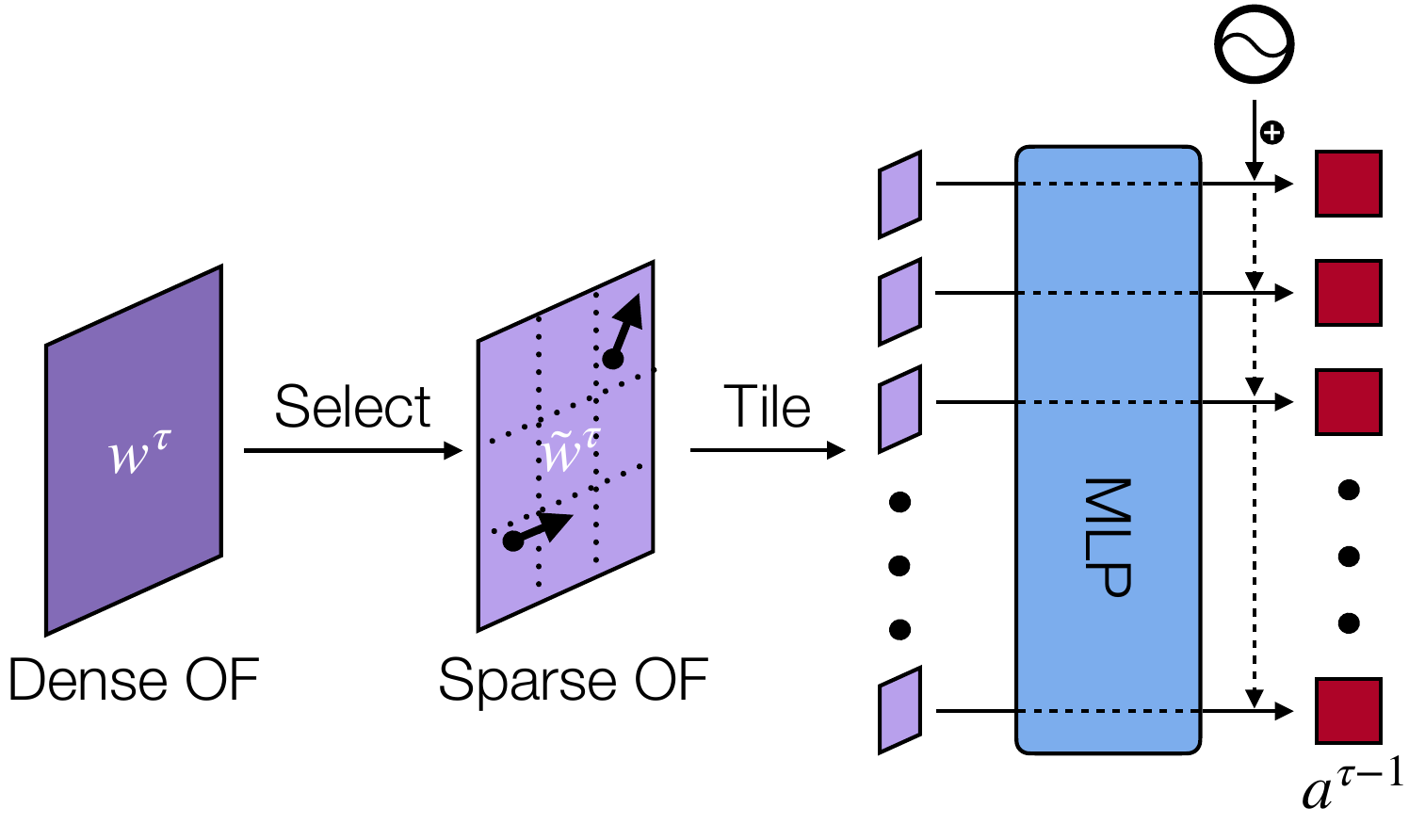}
    \caption{Sparse optical flow (OF) encoder. The dense OF is sampled and tiled in a $16\times 16$ grid. Each OF tile is fed independently to the MLP, and then combined with a learnable positional encoding into a code.}
    \label{fig:control}
\end{figure}

To condition the video generation on the selected optical flow vectors, we introduce an encoding procedure. First, we construct a binary mask $m \in \{0, 1\}^{1 \times H \times W}$ such that $m_{ij} = 1$, $\forall (i, j) \in {\cal S}$ and $m_{ij} = 0$ otherwise. This mask is concatenated in the channel dimension with $m \odot w^\tau$ to form a tensor $\tilde w^\tau$ of shape $(3, H, W)$, which we refer to as the \emph{sparse optical flow}. $\tilde w^\tau$ is further tiled into a $16 \times 16$ grid. Each of these tiles is independently projected through an MLP and augmented with a learnable positional encoding to output a code (see Figure~\ref{fig:control}). This particular design of the optical flow encoder is a trade-off between having a restricted receptive field (because each tile is processed independently) and efficiency. A small receptive field is needed to ensure that separate controls minimally interact before being fed to the cross-attention layers. We found that this is crucial to enable the independent manipulation of objects.

\section{Experiments}

In this section, we evaluate \methodName to see how controllable the video generation is, \emph{i.e.}, how much the generated object motion correlates with the input motion control (see next section), and to assess the image and sequence quality on three datasets with different scene and texture complexities, as well as different object dynamics. We report several metrics, such as FVD~\cite{Unterthiner2018TowardsAG} and average LPIPS~\cite{zhang2018unreasonable}, PSNR, SSIM~\cite{Wang2004ImageQA} and FID~\cite{Parmar2022OnAR}.
Implementation and training details are in the supplementary material.

\begin{figure}[t]
    \centering
    \includegraphics[width=\linewidth]{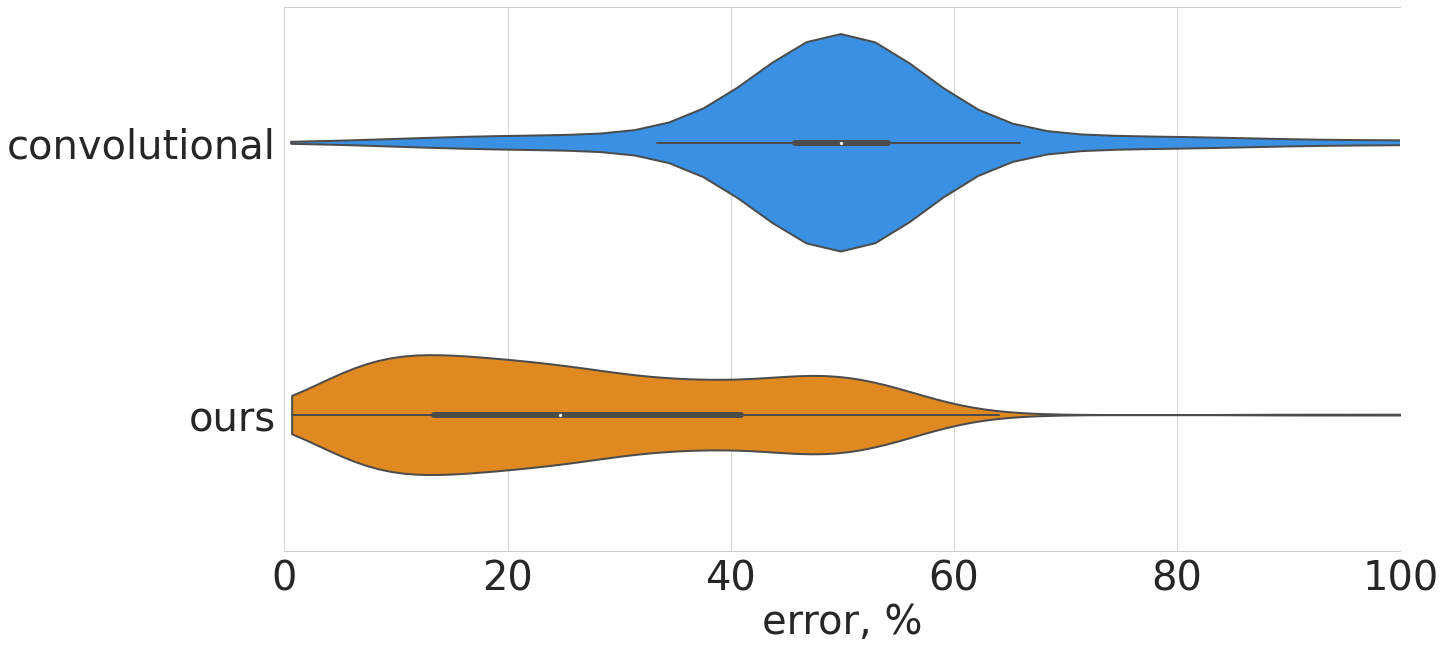}
    \caption{Violin plots of the local error distributions. Top: the distribution of the errors when using a convolutional encoder of the motion controls (\emph{i.e.}, with a large receptive field). Bottom: the distribution with our encoder (\emph{i.e.}, with a limited receptive field). Our encoder not only reduces the overall average error but also tends to have more errors of smaller magnitude when compared to the convolutional encoder. In contrast, the convolutional encoder shows a distinct long tail in errors of larger magnitude.
    }
    \label{fig:encoder}
\end{figure}

\subsection{Evaluation of Intended vs Generated Motion}

The objective of our training is to build a model to generate videos that can be controlled by specifying motion through $a^{\tau-1}$ as a set of shifts at some user-chosen pixels. To evaluate how much the trained model follows the intended control, we introduce the following metrics: Local and global errors.
To compute them, we sample an image from the test videos. Then, we randomly select one object in the scene and apply a random motion control input to a pixel of that object. Because the generated images are of high-quality, we can use a pre-trained optical flow estimation model \cite{teed2020raft} to calculate the optical flow between the first image and the generated one. In principle, one could measure the discrepancy between the input motion shift and the generated one at the same pixel. However, since single pixel measurements are too noisy to use, we assume that all pixels within a small neighborhood around the selected pixel move in the same way, and then average the estimated optical flow within that neighborhood to compare it to the input control vector (depending on what we focus on, we use the relative $L_2$ norm or a cosine similarity). We call this metric the \emph{local error} (see Figure~\ref{fig:nc} on the left in blue).
Notice that this metric is quite coarse.
For example, in some cases the model could use the motion input to rotate an object, instead of translating it. Also, the chosen neighborhood (whose size is fixed) may not fully cover just the object of interest.  
Nonetheless, this metric still provides a useful approximation of the average response of the trained model to the control inputs.
We also assume that the motion generated far away from where the motion control is applied should be zero on average (although in general a local motion could cause another motion far away through a chain reaction). To assess this, we calculate the average $L_2$ norm of the estimated optical flow outside some neighborhood of the controlled pixel. We call this metric the \emph{global error} (see Figure~\ref{fig:nc} on the right in orange). 

\subsection{Datasets}

We evaluate \methodName on the following three datasets:

\noindent\textbf{CLEVRER}~\cite{Yi2020CLEVRERCE} is a dataset containing 10K training and 1000 test videos capturing a synthetic scene with multiple simple objects interacting with each other through collisions. We cropped and downsampled the videos to $128\times 128$ resolution. On CLEVRER we show the ability of our model to model complex cascading interactions and also to learn the control of long-term motions (\emph{i.e.}, motions that once started at one frame can last for several frames).

\noindent\textbf{BAIR}~\cite{Ebert2017SelfSupervisedVP} is a real dataset containing around 44K $256\times 256$ resolution videos of a robot arm pushing toys on a flat square table. The visual complexity of BAIR is much higher than that of CLEVRER. The objects on the table are diverse and include non-rigid objects, such as stuffed toys, that have different physical properties. However, in contrast to CLEVRER, its interactions are simpler and do not require modeling long-term dynamics.

\noindent\textbf{iPER}~\cite{liu2019liquid} captures 30 humans with diverse styles performing various movements of varying complexity. The official train/test split separates the dataset into 164 training and 42 test clips. Although our main focus is to work with multi-object datasets, we use this dataset for two reasons: 1) We can test how \methodName learns to control articulated objects, such as humans; 2) We can compare to the related work iPOKE~\cite{blattmann2021ipoke}, which has already been tested on this dataset (and is not designed for multi-object datasets).

\subsection{Ablations}


\noindent\textbf{Sparse optical flow encoder.} First, we show the importance of using a sparse optical flow encoder with a restricted receptive field. We observe that such an encoder is essential for the model to learn to independently control different objects in the scene. We manually annotated 128 images from the test set of the BAIR dataset. For each image we store a list of pixel coordinates that belong to objects in the scene, 1-2 pixels per object, 3 pixels for the robot arm. We use the local error to compare our encoder with a convolutional sparse optical flow encoder from \cite{blattmann2021ipoke}. In Figure~\ref{fig:encoder}, we show that our restricted receptive field optical flow encoder outperforms the convolutional one. For qualitative comparisons, see supplementary materials.

\begin{figure}[t]
    \centering
    \includegraphics[width=\linewidth, trim=0cm 0cm 0cm 0cm, clip]{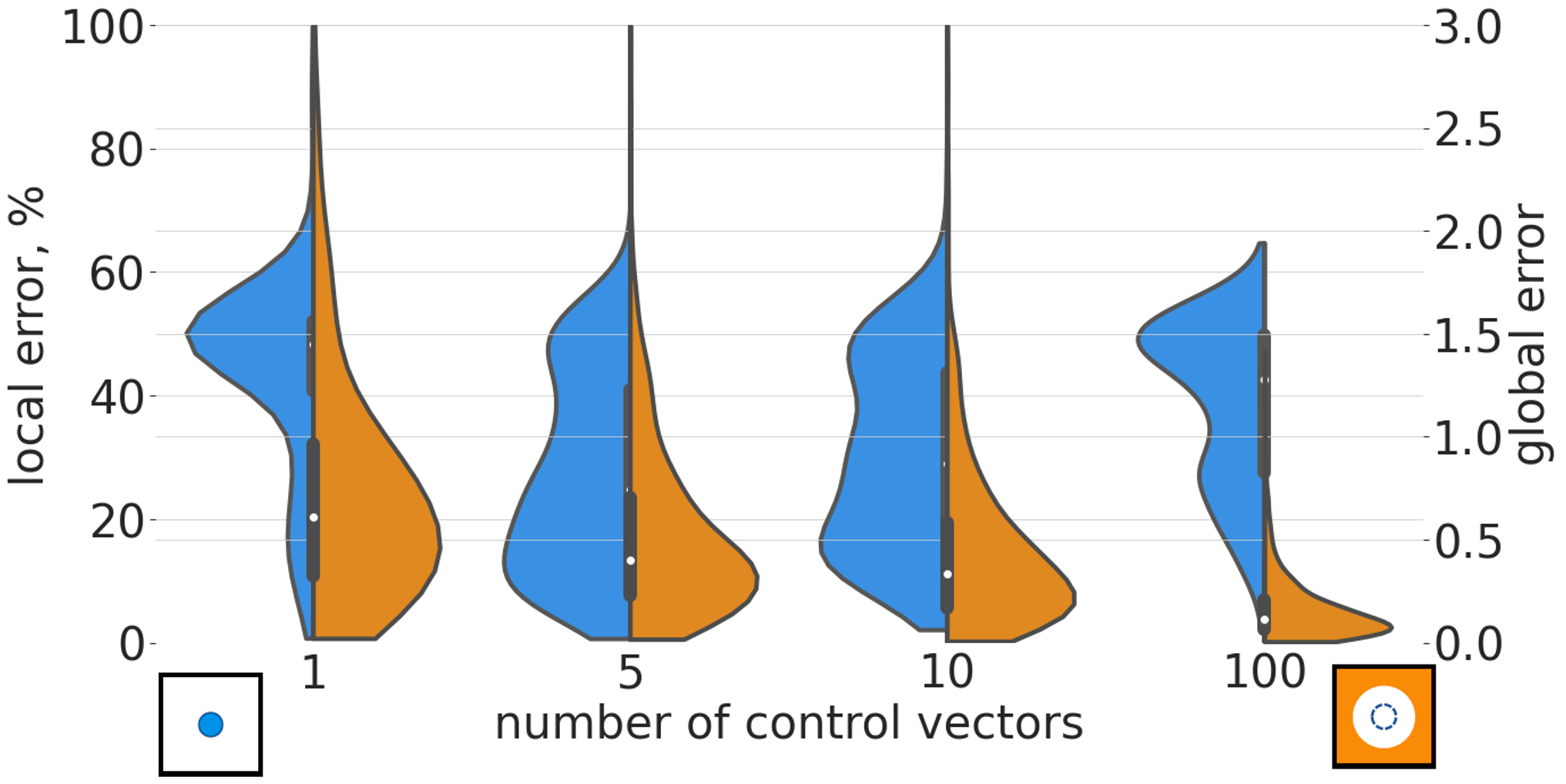}
    \caption{The effect of the number of control vectors when training on the \emph{BAIR} dataset. The \emph{local error} (in blue and on the left-hand side) is the error of the optical flow in the neighborhood of the controlled pixel, while the \emph{global error} (in orange and on the right-hand side) is the average optical flow vector outside of a circle around the interacted pixel. A smaller local error indicates a better response of the model to the control, while a low global error ensures that only the object of interest moves. The two boxes at the bottom of the figure show the regions across which the error is averaged (left - local error, right - global error). Best viewed in color.}
    \label{fig:nc}
\end{figure}

\begin{table}[t]
    \centering
    \footnotesize
    \begin{tabular*}{\linewidth}{@{\extracolsep{\fill}}ll|ccccc@{\extracolsep{\fill}}}
        \toprule
            $n_c$ & $\pi$ & LPIPS$\downarrow$ & PSNR$\uparrow$ & SSIM$\uparrow$ & FID$\downarrow$ & FVD$\downarrow$\\
        \hline
            5 & 0.0 & 0.288 & 24.88 & 0.86 & 9.9 & 401 \\
            5 & 0.25 & \textbf{0.123} & \textbf{30.11} & \textbf{0.93} & \underline{6.0} & 86 \\
            5 & 0.5 & \underline{0.126} & \underline{30.07} & \textbf{0.93} & \textbf{5.7} & \textbf{70} \\
            5 & 0.75 & 0.133 & 29.41 & \underline{0.92} & 6.1 & \underline{83} \\
            \hline
            1 & 0.5 & 0.144 & 28.78 & 0.92 & 6.3 & 83 \\
        \bottomrule
    \end{tabular*}
    \caption{Evaluation of the generated videos on \emph{CLEVRER} under different input sparsity ($n_c$) and randomization ($\pi$).}
    \label{tab:clevrer}
\end{table}

\noindent\textbf{Randomized conditioning.} In Table~\ref{tab:clevrer}, we demonstrate the importance of our randomized conditioning scheme, where we randomly switch off the conditioning with respect to both the past frame and the control input with probability $\pi$. The comparisons on the CLEVRER dataset show that randomized conditioning is crucial for the single frame quality as well as for temporal consistency (see the FVD metric).

\noindent\textbf{Number of control inputs during the training.} $n_c$ plays an important role in enabling the independent control over separate objects. In principle, responses to the control should be more and more decorrelated as we increase the number of control inputs.
However, values of $n_c$ that are too large would increase the gap between the training and the test conditions, where often only 1 control is used. At the same time, using fewer optical flow vectors during the training stimulates the network to learn interactions (\emph{i.e.}, correlations with other objects). Therefore, we observe a trade-off between controllability and learning correct dynamics in the choice of $n_c$ (see Figures~\ref{fig:nc} and \ref{fig:nc_vis}). We choose $n_c=5$ in the remaining experiments.

\begin{figure}[t]
    \centering
    \begin{tabular}{@{}r@{\hspace{0.5mm}}c@{\hspace{0.5mm}}c@{\hspace{0.5mm}}c@{}}
        \rotatebox{90}{\makebox[2.7cm][c]{$n_c = 100$}} & 
        \animategraphics[width=2.7cm, autoplay, loop]{3}{Figures/bair_robot_n100/image_}{0}{5} &
        \animategraphics[width=2.7cm, autoplay, loop]{3}{Figures/bair_zebra_n100_m/image_}{0}{6} &
        \animategraphics[width=2.7cm, autoplay, loop]{3}{Figures/bair_zebra_back_n100_m/image_}{0}{3} \\
        \rotatebox{90}{\makebox[2.7cm][c]{$n_c = 1$}} & 
        \animategraphics[width=2.7cm, autoplay, loop]{3}{Figures/bair_robot_n1/image_}{0}{5} &
        \animategraphics[width=2.7cm, autoplay, loop]{3}{Figures/bair_zebra_n1/image_}{0}{3} &
        \animategraphics[width=2.7cm, autoplay, loop]{3}{Figures/bair_zebra_back_n1/image_}{0}{3} \\
        \rotatebox{90}{\makebox[2.7cm][c]{$n_c = 5$}} & 
        \animategraphics[width=2.7cm, autoplay, loop]{3}{Figures/bair_robot_n5/image_}{0}{5} &
        \animategraphics[width=2.7cm, autoplay, loop]{3}{Figures/bair_zebra_n5/image_}{0}{3} &
        \animategraphics[width=2.7cm, autoplay, loop]{3}{Figures/bair_zebra_back_n5/image_}{0}{3} \\
    \end{tabular}
    \caption{Ablation of the number of control vectors $n_c$ during training. The videos in each column start from the same initial frame and are generated with the same sequence of control vectors. Notice, however, that for $n_c = 100$ one has to use more controls at inference to bridge the gap between training and test settings. With too many control vectors during training the model demonstrates decent control over background objects, but struggles modelling interactions. With too few control vectors the interactions are modelled well, while the model lacks control over background objects. With the optimal $n_c = 5$ we get the best of the two worlds. Use Acrobat Reader to play the videos.}
    \label{fig:nc_vis}
\end{figure}

\subsection{Quantitative Results}


\noindent\textbf{Realism and motion consistency.} Following \cite{menapace2021playable}, we train \methodName on BAIR and then autoregressively generate 29 frames given the first frame and a set of controls at each generation step, decoded from the ground truth videos (in our case from the corresponding optical flows). We report the metrics for 1 and 5 control vectors at each timestamp and show that the \textit{5 controls} version outperforms all prior work (see Table~\ref{tab:bair}). Moreover, notice that prior work on controllable generation on BAIR only focuses on modeling the actions of the robot arm, while \methodName is able to also effectively control other objects in the scene.
\begin{table}[t]
    \footnotesize
    \begin{tabularx}{\linewidth}{@{}Xcrr@{}}
        \toprule
            Method & LPIPS$\downarrow$ & FID$\downarrow$ & FVD$\downarrow$\\
        \hline
            MoCoGAN~\cite{tulyakov2018mocogan} & 0.466 & 198.0 & 1380 \\
            MoCoGAN+~\cite{menapace2021playable} & 0.201 & 66.1 & 849 \\
            SAVP~\cite{lee2018stochastic} & 0.433 & 220.0 & 1720 \\
            SAVP+~\cite{menapace2021playable} & 0.154 & 27.2 & 303 \\
            CADDY~\cite{menapace2021playable} & 0.202 & 35.9 & 423 \\
            \cite{huang2022layered} \emph{positional}  & 0.202 & 28.5 & 333 \\
            \cite{huang2022layered} \emph{affine} & 0.201 & 30.1 & \underline{292} \\
            \cite{huang2022layered} \emph{non-param} & 0.176 & 29.3 & 293 \\
            GLASS~\cite{davtyan2022glass} & \underline{0.118} & \underline{18.7} & 411 \\
        \hline
            \methodName (ours) \textit{5 controls} & \textbf{0.112} & \textbf{18.2} & \textbf{264} \\
            \methodName (ours) \textit{1 control} & 0.142 & 19.2 & 339 \\
        \bottomrule
    \end{tabularx}
    \centering
    \caption{Evaluation on the \emph{BAIR} dataset.}
    \label{tab:bair}
\end{table}
We also compare our model on the benchmark introduced in \cite{blattmann2021ipoke} and generate 9 frames starting from a single initial frame. Although our model does not outperform \cite{blattmann2021ipoke}, it does better than all the other prior work, as indicated by our FVD metric (see Table~\ref{tab:iper}). One should also notice that \methodName does not make use of the same information used in iPOKE \cite{blattmann2021ipoke} (\emph{e.g.}, what a background is).

\begin{table}[t]
    \centering
    \footnotesize
    \begin{tabular*}{\linewidth}{@{\extracolsep{\fill}}l@{\hspace{.3cm}}r@{\hspace{.25cm}}c@{\hspace{.25cm}}c@{\extracolsep{\fill}}}
        \toprule
            Method & FVD$\downarrow$ & LPIPS$\downarrow$ &  SSIM$\uparrow$\\
        \hline
            Hao~\cite{hao2018controllable} & 235.08 & 0.11 & 0.88 \\
            Hao~\cite{blattmann2021ipoke} \textit{w/ KP} & 141.07 & \textbf{0.04} & \textbf{0.93} \\
            II2V~\cite{blattmann2021understanding} & 220.34 & 0.07 & \underline{0.89} \\
            iPOKE~\cite{blattmann2021ipoke} & \textbf{77.50} & \underline{0.06} & 0.87 \\
        \hline
            \methodName (ours) \textit{1st frame} & \underline{133.08} & 0.09 & 0.86 \\
            \methodName (ours) \textit{all frames} & 134.35 & 0.09 & 0.87 \\
        \bottomrule
    \end{tabular*}
    \caption{Evaluation on the \emph{iPER} dataset.}
    \label{tab:iper}
\end{table}

\begin{figure}[t]
    \centering
    \includegraphics[width=\linewidth]{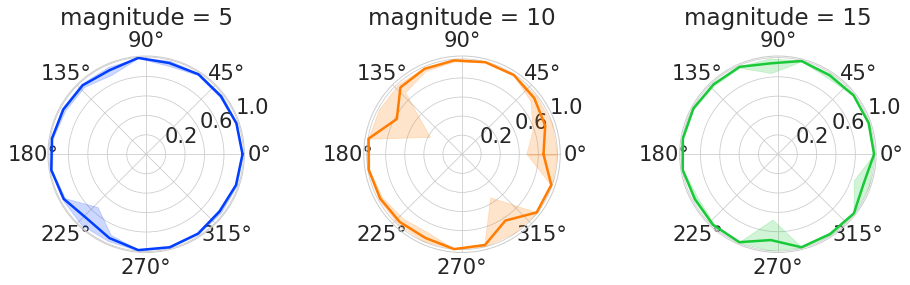}
    \caption{Average cosine error and 95\% confidence intervals between the control vector and the resulting optical flow  depending on the magnitude of the control (in pixels) and its direction on CLEVRER. 
    }
    \label{fig:polar}
\end{figure}

\noindent\textbf{Scene dynamics and interactions.} We chose to assess how well \methodName models the physics of the scene and object interactions on the CLEVRER dataset. For this purpose, we generate 15 frames starting from a single frame and a set of control vectors. This time we do not specify future controls like in BAIR and let the model simulate the learned physics. We repeat the experiment with 1 and 5 control vectors. The metrics in Table~\ref{tab:clevrer} show that \methodName models the interior object dynamics well, which is further supported by the qualitative results in section~\ref{sec:qual}.\\
\noindent\textbf{Controllability.} We asses how robust \methodName is to changes in the control parameters, such as the direction and the magnitude of the control vectors. We manually annotate 128 images from the CLEVRER dataset by indicating 1 potential control point per object. We then sample some points and random control vectors from the annotated ones and feed those to the model to generate the next frame. We calculate the local error with the cosine distance between the normalized optical flow vectors. We show how this metric changes with the parameters of the control in Figure~\ref{fig:polar}.

Additionally, we evaluate controllability  against the baselines on BAIR using the local error. However, the baselines often have different degrees of controllability as well as the nature of the control inputs. For instance, CADDY~\cite{menapace2021playable} conditions on a discrete action label that is inferred from 2 consecutive frames. The global action of GLASS~\cite{davtyan2022glass} in turn is very similar to our optical flow control, but allows only for a single optical flow vector that is applied to the main object in the scene. Therefore, for this evaluation we focus on the main object in the scene, the robotic arm. In Table~\ref{tab:control_baselines} we report the local error under 3 scenarios: 1) pseudo-ground truth (GT) control inferred from the known future frame 2) with a randomized intended control, with a uniformly sampled direction and a norm sampled from from ${\cal N}(7.19, 5.12)$, which approximates the distribution of the magnitude of the control in the GT data; 3) out of distribution randomized control with a norm sampled from ${\cal N}(10, 0.1)$ (o.o.d. S), or ${\cal N}(20, 0.1)$ (o.o.d. L). One can see that \methodName is not only more controllable than in prior work, but also more robust to distribution shifts in the control.

\begin{figure}[t]
    \centering
    \begin{tabular}{@{}c@{\hspace{0.3mm}}c@{\hspace{0.3mm}}c@{\hspace{0.3mm}}c@{}}
    {\animategraphics[width=2cm, autoplay, loop]{3}{Figures/cl_slow/image_}{0}{9}} &
    {\animategraphics[width=2cm, autoplay, loop]{3}{Figures/cl_fast/image_}{0}{9}} &
    {\animategraphics[width=2cm, autoplay, loop]{3}{Figures/cl_other/image_}{0}{9}} &
    {\animategraphics[width=2cm, autoplay, loop]{3}{Figures/cl_together/image_}{0}{9}} \\
    \end{tabular}
    \caption{In each column \methodName is given the controls indicated as red arrows. Starting from the same frame, \methodName generates diverse videos, following accurately the magnitude (1st and 2nd videos) and direction (1st to 3rd videos) of the motion control input, as well as multiple inputs  (4th video). Notice that since \methodName has memory and was trained also to generate frames without control, it can learn to propagate motion across many frames from a single initial control input. To play videos, use Acrobat Reader.
    }
    \label{fig:clevrer}
\end{figure}

\begin{table}[!t]
    \centering
    \footnotesize
    \begin{tabularx}{\linewidth}{@{}l|c@{\hspace{.4cm}}c@{\hspace{.4cm}}c@{\hspace{.4cm}}c@{}}
    \toprule
         & pseudo-GT & from GT dist. & o.o.d. S & o.o.d. L \\
        \hline
        CADDY & 6.29 & - & - & - \\
        GLASS & 1.75 & 25.00 & 25.07 & 30.06 \\
        YODA (ours) & \textbf{1.41} & \textbf{1.77} & \textbf{2.01} & \textbf{3.37} \\
    \bottomrule
    \end{tabularx}
    \caption{Local error in pixels of the control applied to the robot arm. Average of 5 runs is reported.}
    \label{tab:control_baselines}
\end{table}

\subsection{Qualitative Results}\label{sec:qual}

In this section we provide some visual examples of the generated sequence. On the BAIR dataset, we highlight the capability of \methodName to move, rotate and deform separate objects without the robot arm physically touching them. Figure~\ref{fig:bair_teaser} shows different object manipulations on the BAIR test set. 
Notice how the model has learned also the 3D representations and interactions of the objects.
Figure~\ref{fig:clevrer} shows the diversity of generated videos that share the same initial frame, but use different control signals. Notice the high correlation between the intended and generated motions and the ability of the model to correctly predict the interactions between the colliding objects. 
More qualitative results, including sequences on the iPER and CLEVERER datasets, as well as  controllability evaluation and possible applications of \methodName, are in the supplementary materials.

\section{Conclusion}

In this paper we introduced \methodName, a novel method for controllable video generation from sparse motion input. Our experimental evaluation demonstrates the ability of \methodName to generate realistic multi-object videos, which involves learning the extents and interactions between multiple objects despite only passively observing their correlated motions.

\noindent\textbf{Acknowledgements}
This work was supported by grant 188690 of the Swiss National Science Foundation.

\appendix

\section{Appendix}

In the main paper we have introduced a novel method for controllable video generation of multi-object scenes, which we call \methodName. This supplementary material provides details that could not be included in the main paper. In section~\ref{sec:arch} we explain details on the architecture and the training of our model. Section~\ref{sec:app_qual} provides more qualitative results with \methodName, including more control scenarios on the BAIR~\cite{Ebert2017SelfSupervisedVP} dataset and longer video generation results. Section~\ref{sec:demo} introduces our interactive demo. Section~\ref{sec:app} provides discussion on possible applications of our method, such as object segmentation.

\section{Architecture and training details}\label{sec:arch}

\noindent\textbf{Autoencoder.} We trained a VQGAN~\cite{esser2021taming} autoencoder per dataset, using the official \texttt{taming transformers} repository.\footnote{\url{https://github.com/CompVis/taming-transformers}} 

Here we provide some preliminaries for VQGAN.  VQGAN consists of an encoder $E$ and a decoder $G$ networks that are jointly trained to reconstruct input images. The encoder maps the image $x \in \mathbb{R}^{3 \times H \times W}$ into a compressed latent representation $z \in \mathbb{R}^{n_z \times h \times w}$. This representation is quantized to a discrete set of learned embeddings $Z = \{z_k\}_{k = 1}^K, z_k \in \mathbb{R}^{n_z}$ to obtain $z_q \in \mathbb{R}^{n_z \times h \times w}$. The quantization is done by selecting the closest element from $Z$ for each of $h \times w$ vectors in $z$. $z_q$ is then fed to the decoder network to get $\hat x$ - the reconstructed image. The networks as well as the codebook $Z$ are trained to optimize the following loss:

\begin{equation}
\begin{split}
    {\cal L}_{\text{VQ}}(E, G, Z) = \|x -  \hat x\|^2 &+ \|\text{sg}[E(x)] - z_q \|^2 \\ &+ \|\text{sg}[z_q] - E(x)\|^2,
\end{split}
\end{equation}

where $\text{sg}$ denotes the stop-gradient operation.

In addition to this, an adversarial loss can be utilized to enhance the perceptual richness of the codebook. To this end, a discriminator network $D$ is trained to distinguish the reconstructed images from the real ones, while $G$ is trained to fool the discriminator. This can be done via optimizing the vanilla GAN loss~\cite{Goodfellow2014GenerativeAN} in a min/max manner:

\begin{equation}
    {\cal L}_{\text{GAN}}(E, G, Z, D) = \log D(x) + \log(1 - D(\hat x))
\end{equation}

For more details, please, check the original paper~\cite{esser2021taming}.

The configurations of the VQGANs trained in this work can be found in Table~\ref{tab:vqgan}. Notice that we did not use a discriminator for the CLEVRER~\cite{Yi2020CLEVRERCE} dataset.

\noindent\textbf{Sparse optical flow encoder.} The $16\times 16$ tiled grid of sparse optical flow inputs is reshaped and linearly projected to 256 256-dimensional vectors that are fed into 5 subsequent blocks of (batch normalization~\cite{ioffe2015batch}, fully-connected layer and gelu activation~\cite{hendrycks2016gaussian}). The activation is omitted in the last block. This procedure results into a representation of the control input as 256 256-dimensional vectors.

\begin{figure*}
    \centering
    \includegraphics[width=\linewidth, trim=2cm 6.2cm 2cm 9.2cm, clip]{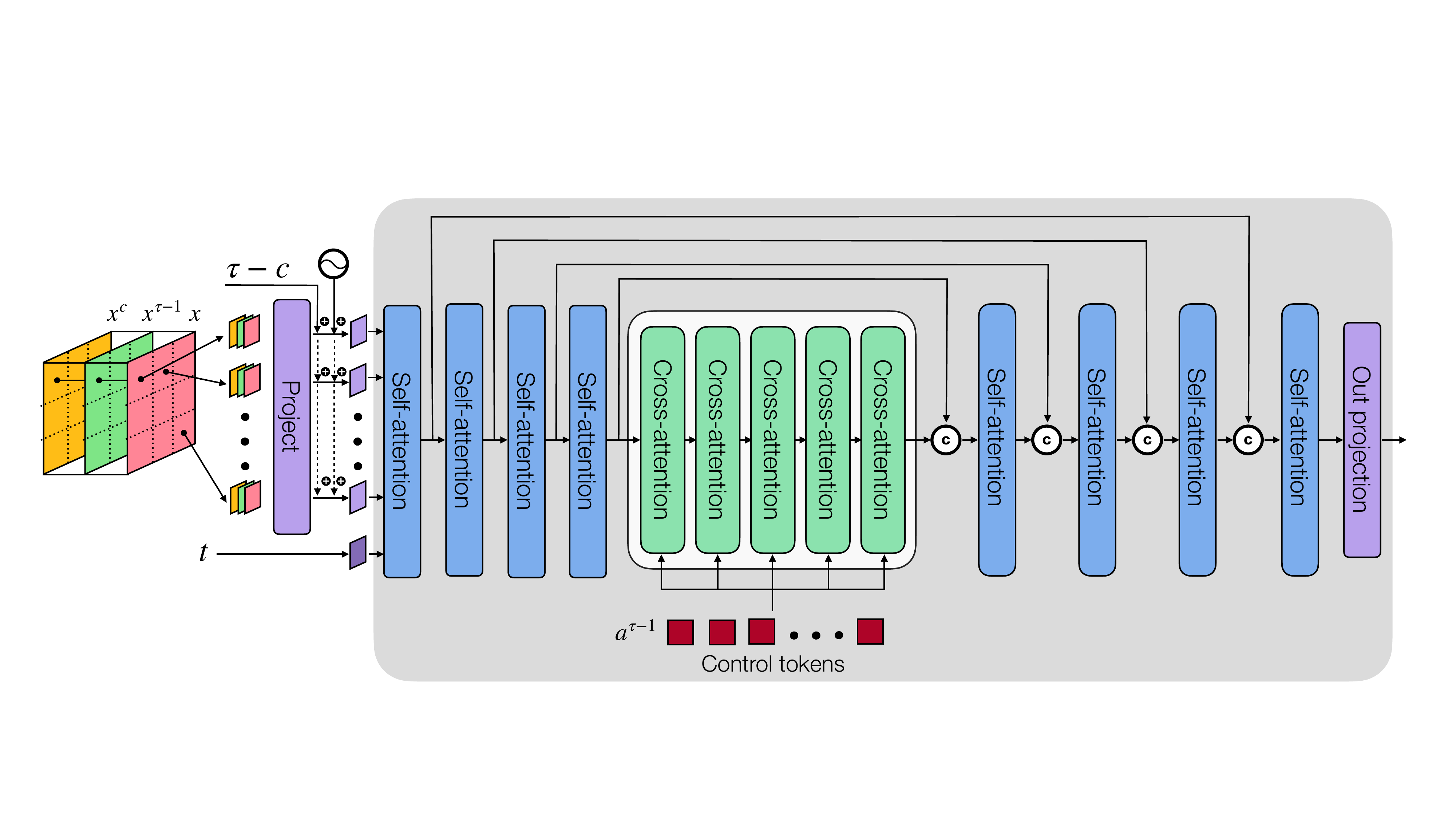}
    \caption{Overview of \methodName's architecture. The noisy target frame $x$, the reference frame $x^{\tau - 1}$ and the context frame $x^c$ are concatenated, reshaped and projected to form a sequence of visual tokens that is augmented with position encodings and embedded relative temporal distance ($\tau-c$) between frames and fed to the U-ViT~\cite{bao2022all} alongside with the embedded time token ($t$). A sequence of control tokens is fused into the pipeline via cross-attention in the bottleneck of the network.}
    \label{fig:force}
\end{figure*}

\noindent\textbf{Main body.} The architecture of the main body of \methodName is adopted from the U-ViT~\cite{bao2022all} architecture of RIVER~\cite{Davtyan_2023_ICCV} with 8 outer self-attention blocks connected through residual concatenation and projection layers and 5 cross-attention blocks in the bottleneck that condition the model on the sequence of control tokens (see Figure~\ref{fig:force}).

\noindent\textbf{Training.} All models are trained for 300K iterations with the AdamW~\cite{Loshchilov2019DecoupledWD} optimizer with the base learning rate equal to $10^{-4}$ and weight decay $5\cdot 10^{-6}$. A learning rate linear warmup for 5K iterations is applied as well as a square root decay schedule. The training takes approximately 3 days on 4 Nvidia GeForce RTX 3090 GPUs and requires at maximum 21 GB per GPU memory for BAIR $256\times 256$ and 12 GB per GPU memory for CLEVRER $128\times 128$ with batch size 16.
For the CLEVRER~\cite{Yi2020CLEVRERCE} dataset, following \cite{Davtyan_2023_ICCV}, random color jittering is additionally used to prevent overfitting. For the iPER~\cite{liu2019liquid} dataset, a random horizontal flip and a random time reversal is applied to prevent overfitting to the pose of the human.
As suggested in \cite{Davtyan_2023_ICCV}, we used $\sigma_{\text{min}} = 10^{-7}$ in the flow matching loss~\cite{lipman2022flow} in all the experiments. In the final models, we used $n_c=5$ optical flow vectors for the control input.

\begin{table*}[t]
    \centering
    \begin{tabularx}{\linewidth}{@{}r|@{\hspace{1cm}}c@{\hspace{2.5cm}}c@{\hspace{2.5cm}}c@{}}
    \toprule
         & \makecell{CLEVRER $128\times128$ \\ \cite{Yi2020CLEVRERCE}} & \makecell{BAIR $256\times256$ \\ \cite{Ebert2017SelfSupervisedVP}} & \makecell{iPER $128\times 128$ \\ \cite{liu2019liquid} }\\
    \hline
        embed\textunderscore dim & 4 & 8 & 4 \\
        n\textunderscore embed & 8192 & 16384 & 16384 \\
        double\textunderscore z & False & False & False \\ 
        z\textunderscore channels & 4 & 8 & 4 \\
        resolution & 128 & 256 & 128 \\
        in\textunderscore channels & 3 & 3 & 3 \\
        out\textunderscore ch & 3 & 3 & 3 \\
        ch & 128 & 128 & 128 \\
        ch\textunderscore mult & [1,2,2,4] & [1,1,2,2,4] & [1,2,2,4] \\
        num\textunderscore res\textunderscore blocks & 2 & 2 & 2\\
        attn\textunderscore resolutions & [16] & [16] & [32] \\
        dropout & 0.0 & 0.0 & 0.0 \\
    \hline
        disc\textunderscore conditional & - & False & False \\
        disc\textunderscore in\textunderscore channels & - & 3 & 3 \\
        disc\textunderscore start & - & 20k & 20k \\
        disc\textunderscore weight & - & 0.8 & 0.8 \\
        codebook\textunderscore weight & - & 1.0 & 1.0 \\
    \bottomrule
    \end{tabularx}
    \caption{Configurations of VQGAN~\cite{esser2021taming} for different datasets.}
    \label{tab:vqgan}
\end{table*}

\section{Qualitative results}\label{sec:app_qual}

\begin{figure*}
    \centering
    \begin{tabular}{@{}r@{\hspace{0.5mm}}c@{\hspace{0.5mm}}c@{\hspace{0.5mm}}c@{\hspace{0.5mm}}c@{\hspace{0.5mm}}c@{\hspace{0.5mm}}c@{\hspace{0.5mm}}c@{}}
    \rotatebox{90}{\makebox[2.4cm][c]{Robot motion}} &
    {\animategraphics[width=2.4cm]{3}{Figures/bair_robot_int/image_}{0}{14}} & {\includegraphics[width=2.4cm]{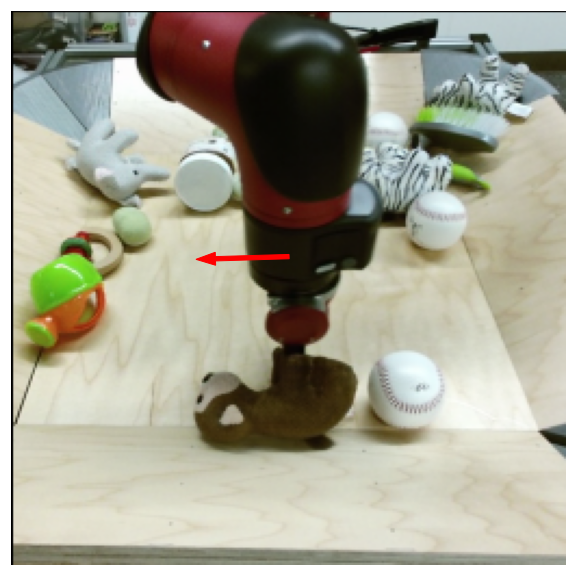}} &
    {\includegraphics[width=2.4cm]{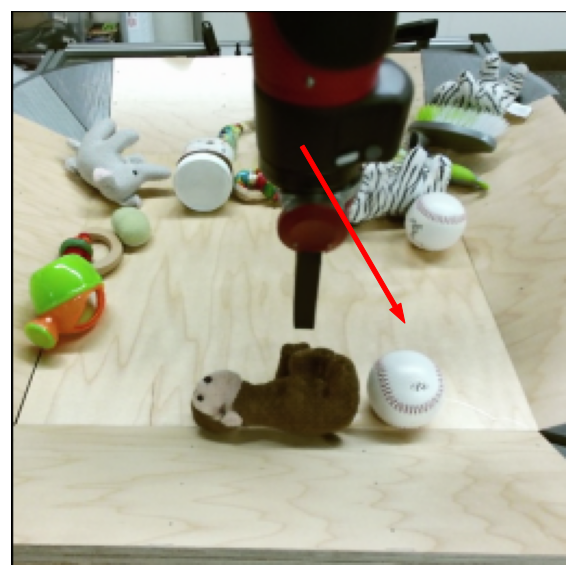}} & 
    {\includegraphics[width=2.4cm]{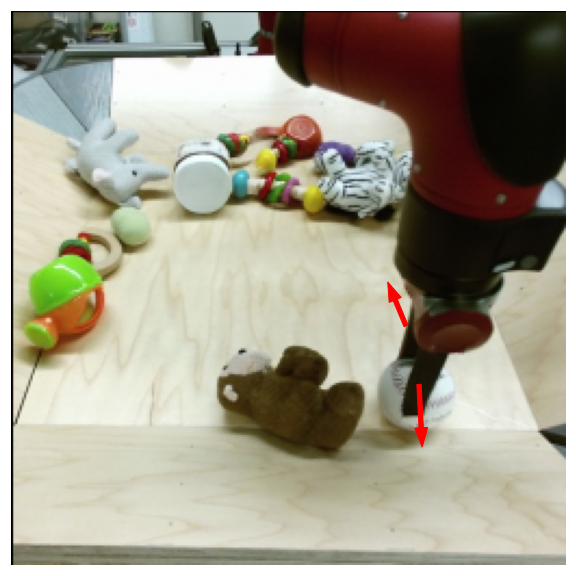}} &
    {\includegraphics[width=2.4cm]{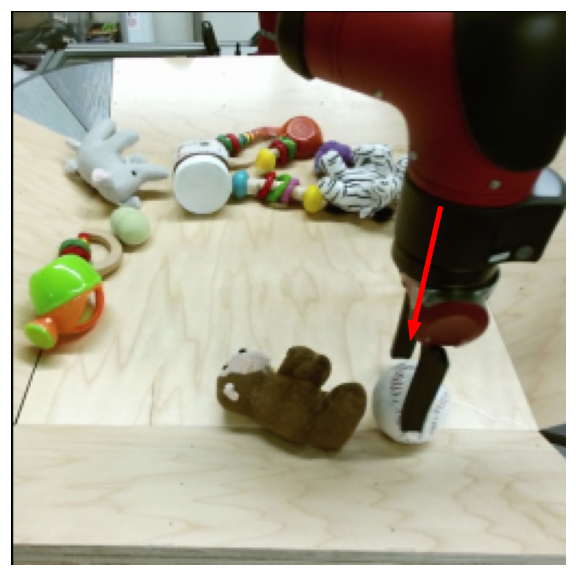}} & 
    {\includegraphics[width=2.4cm]{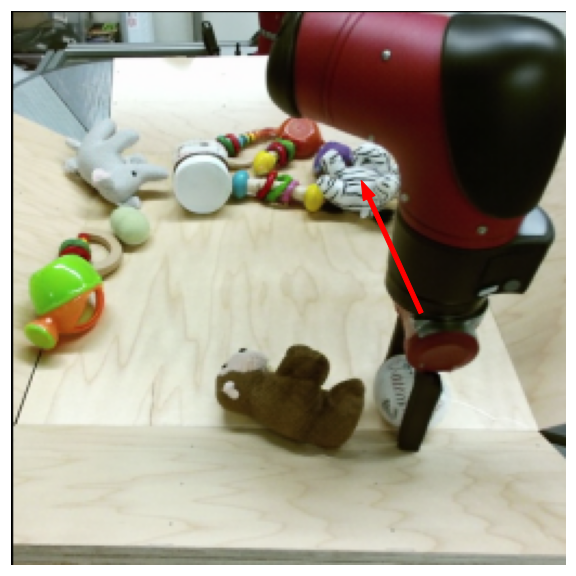}} & 
    {\includegraphics[width=2.4cm]{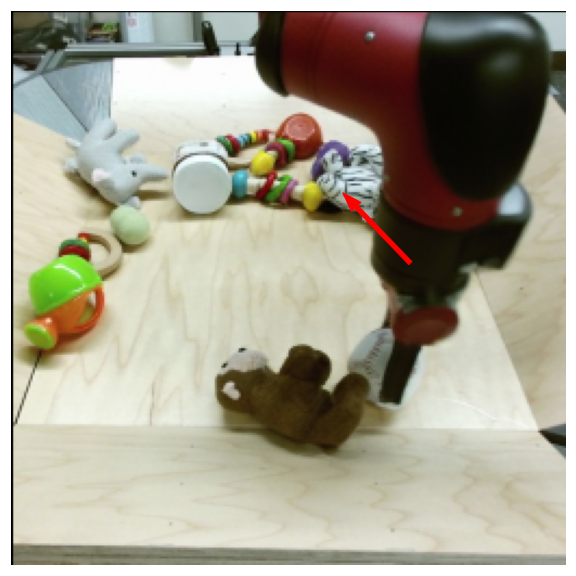}} \\
    \rotatebox{90}{\makebox[2.4cm][c]{Transition}} &
    {\animategraphics[width=2.4cm]{3}{Figures/bair_transition/image_}{0}{13}} & {\includegraphics[width=2.4cm]{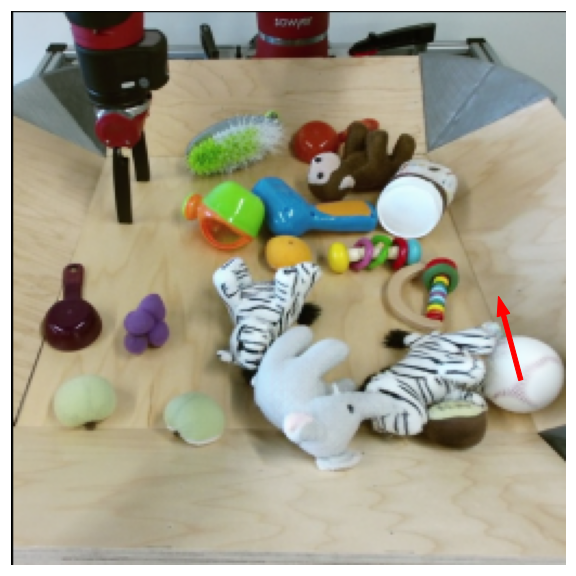}} &
    {\includegraphics[width=2.4cm]{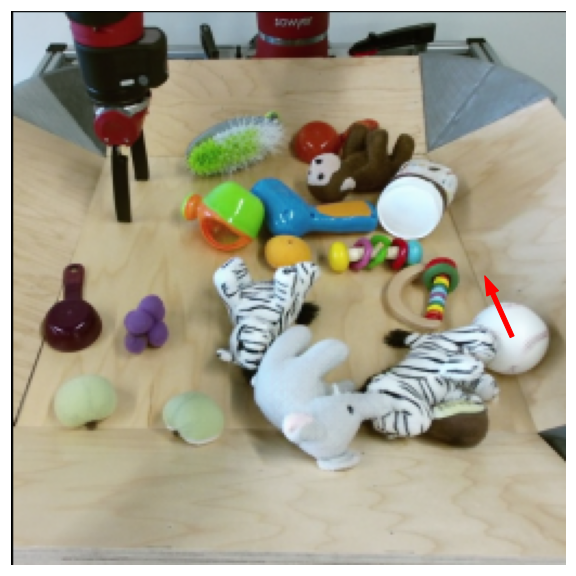}} & 
    {\includegraphics[width=2.4cm]{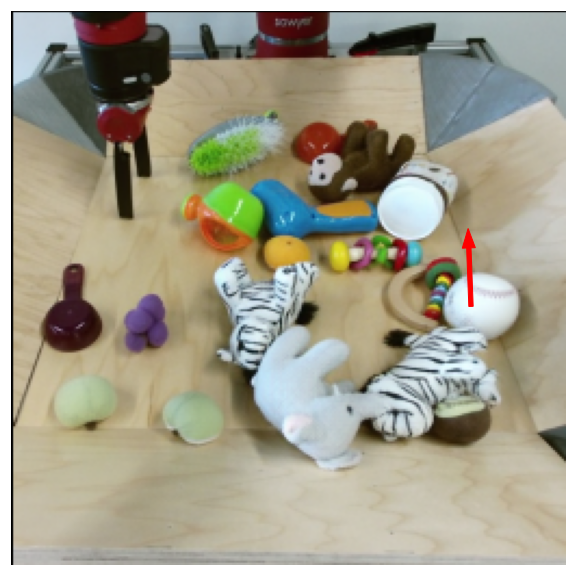}} &
    {\includegraphics[width=2.4cm]{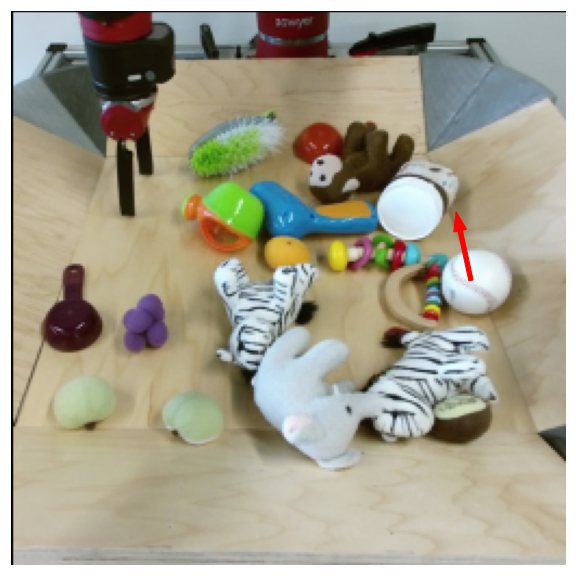}} & 
    {\includegraphics[width=2.4cm]{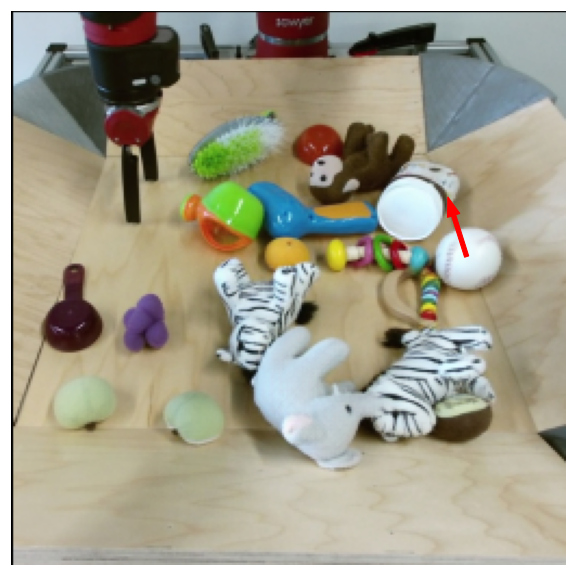}} & 
    {\includegraphics[width=2.4cm]{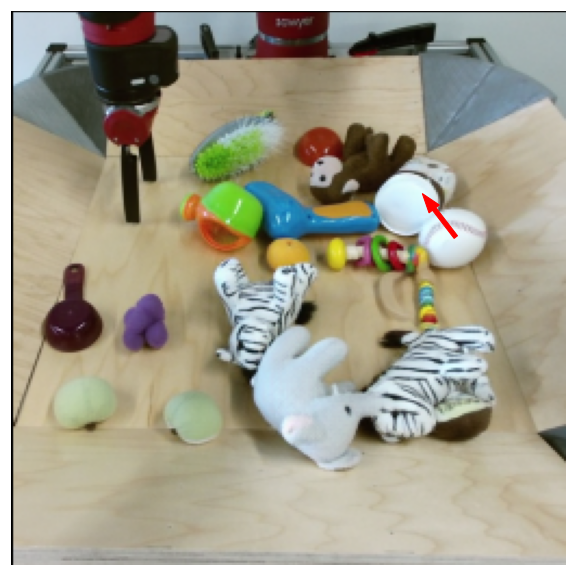}} \\
    \rotatebox{90}{\makebox[2.4cm][c]{Multi-obj. control}} &
    {\animategraphics[width=2.4cm]{3}{Figures/bair_multiobj/image_}{0}{9}} & {\includegraphics[width=2.4cm]{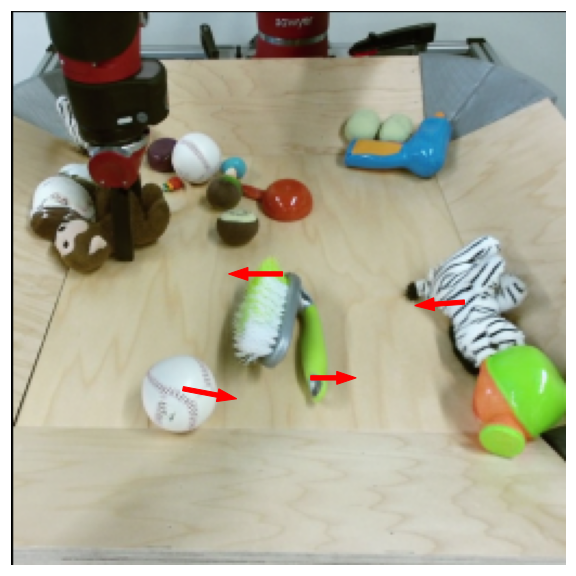}} & 
    {\includegraphics[width=2.4cm]{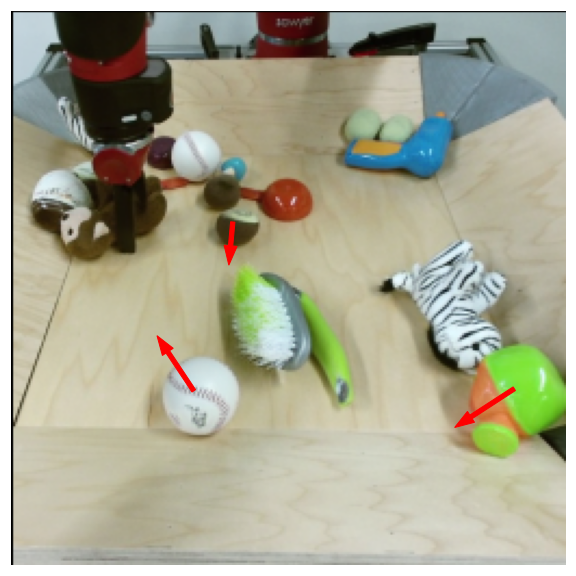}} & 
    {\includegraphics[width=2.4cm]{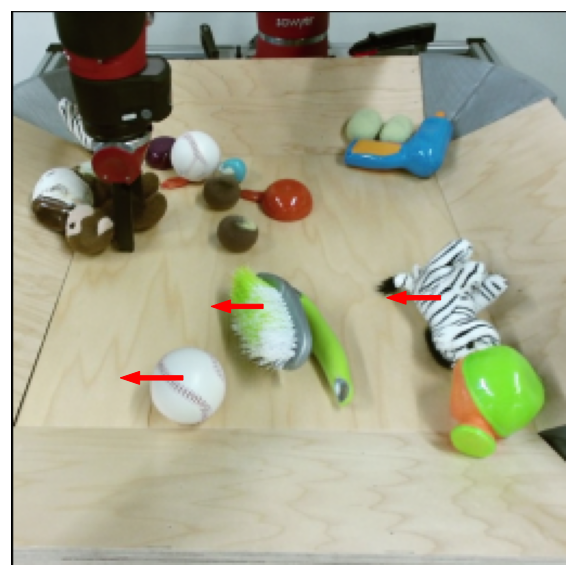}} &
    {\includegraphics[width=2.4cm]{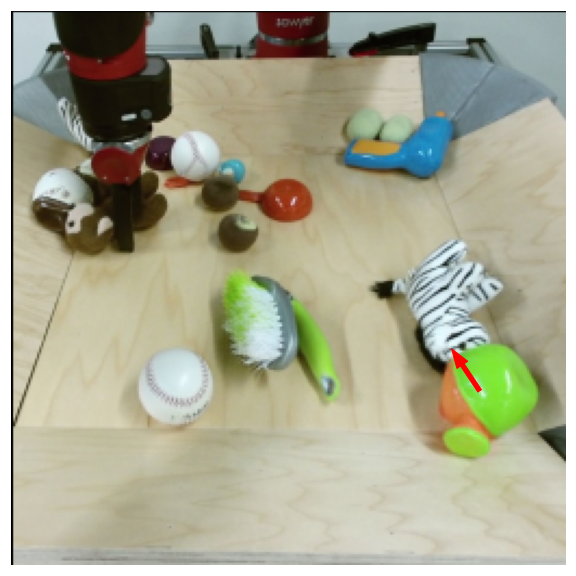}} & 
    {\includegraphics[width=2.4cm]{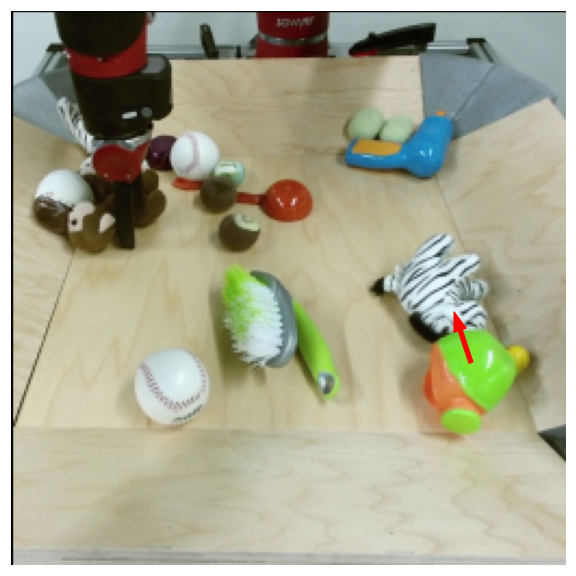}} & 
    {\includegraphics[width=2.4cm]{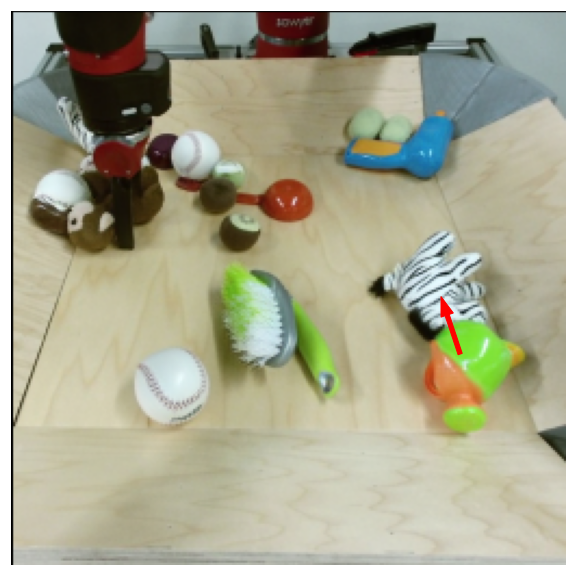}} \\
        \rotatebox{90}{\makebox[2.4cm][c]{Rotation}} & \animategraphics[width=2.5cm]{7}{Figures/supp/bair/rotate_brush/image_}{0}{28} &  \includegraphics[width=2.5cm]{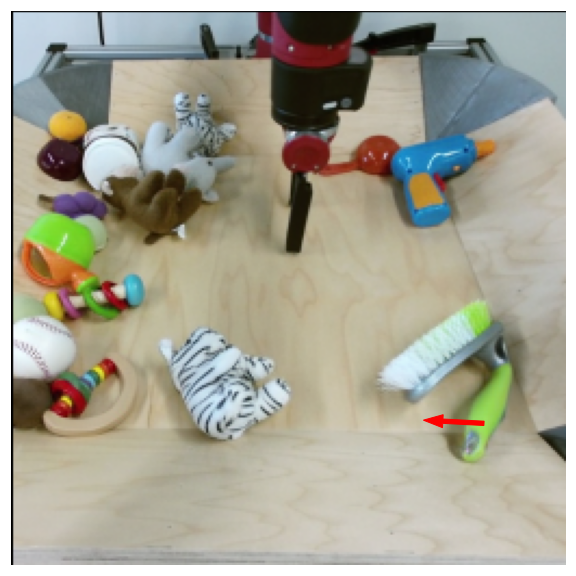} &
        \includegraphics[width=2.5cm]{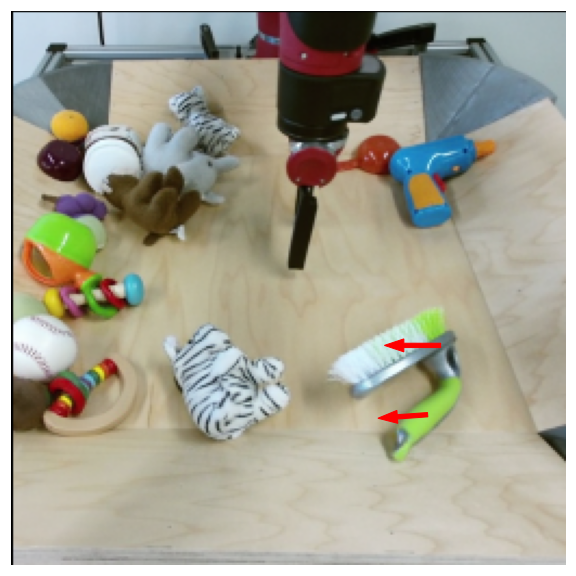} & 
        \includegraphics[width=2.5cm]{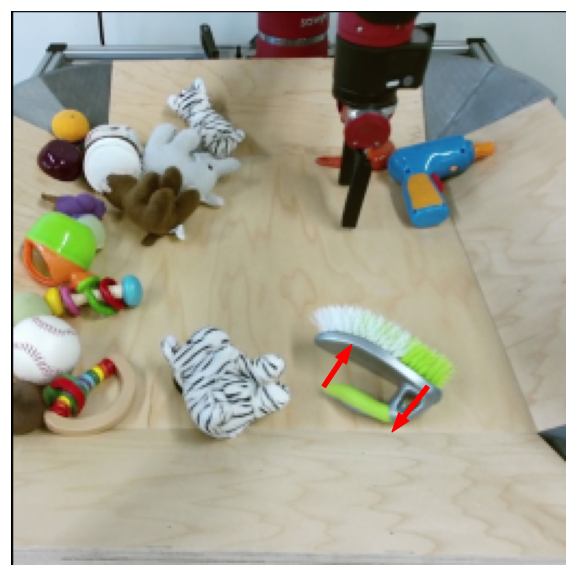} &
        \includegraphics[width=2.5cm]{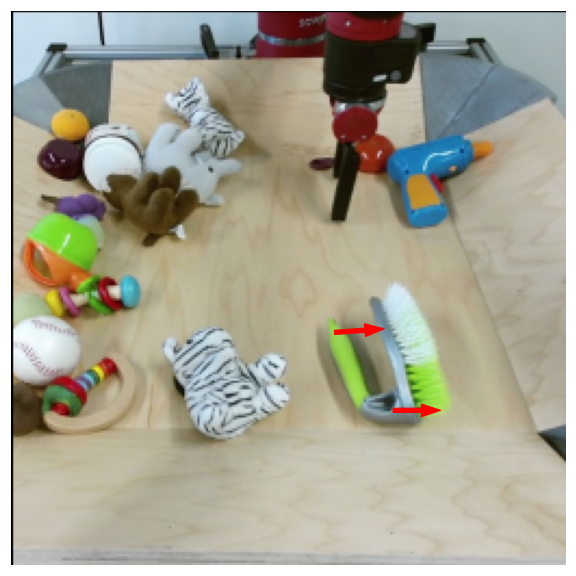} &
        \includegraphics[width=2.5cm]{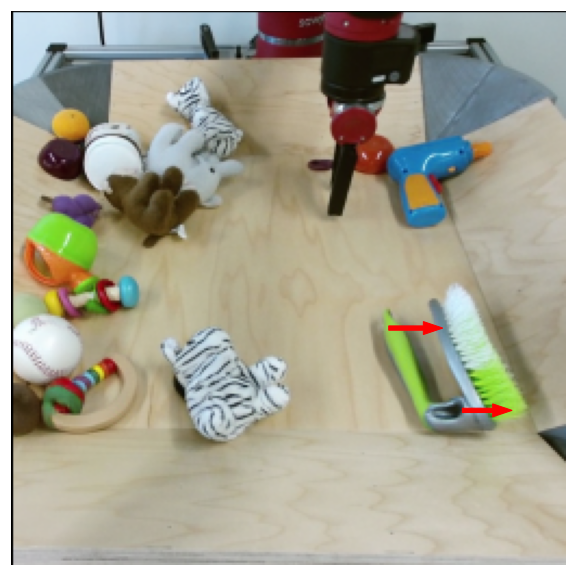} &
        \includegraphics[width=2.5cm]{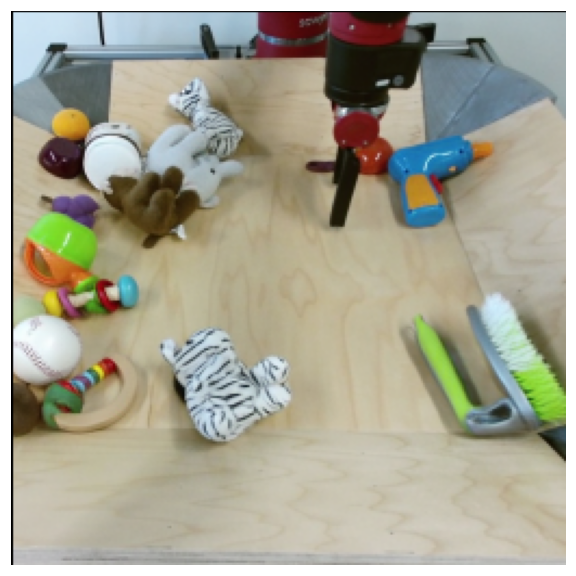}
    \end{tabular}
    \caption{Different scene manipulations with \methodName on the BAIR~\cite{Ebert2017SelfSupervisedVP} dataset. The vectors in red are the control signals specified per frame in an interactive regime. Please, use Acrobat Reader to play the frames in the leftmost column or open the attached \texttt{index.html} file in your browser.}
    \label{fig:bair}
\end{figure*}

\begin{figure*}
    \centering
    \begin{tabular}{@{}c@{\hspace{0.5mm}}c@{\hspace{0.5mm}}c@{\hspace{0.5mm}}c@{\hspace{0.5mm}}c@{\hspace{0.5mm}}c@{\hspace{0.5mm}}c@{\hspace{0.5mm}}c@{}}
        \animategraphics[width=2.5cm]{7}{Figures/supp/clevrer/4/image_}{0}{30} &  \includegraphics[width=2.5cm]{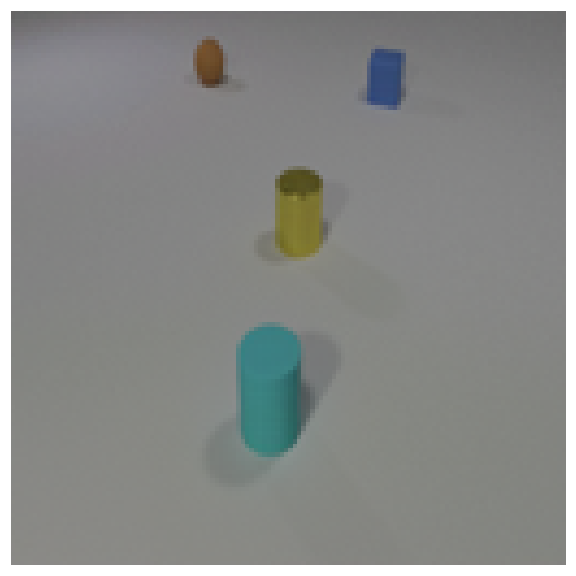} &
        \includegraphics[width=2.5cm]{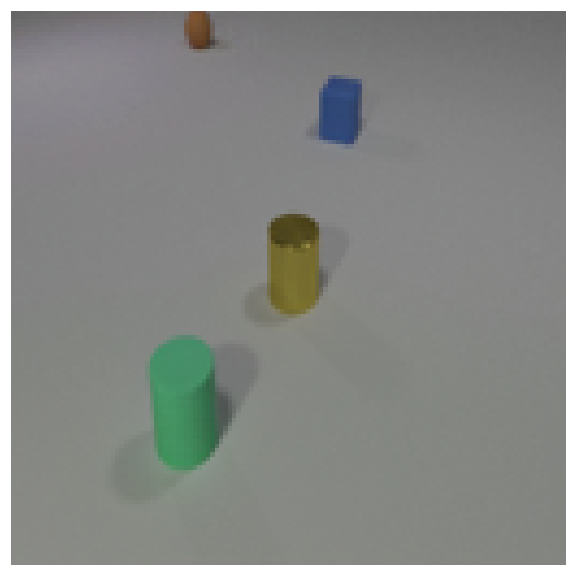} & 
        \includegraphics[width=2.5cm]{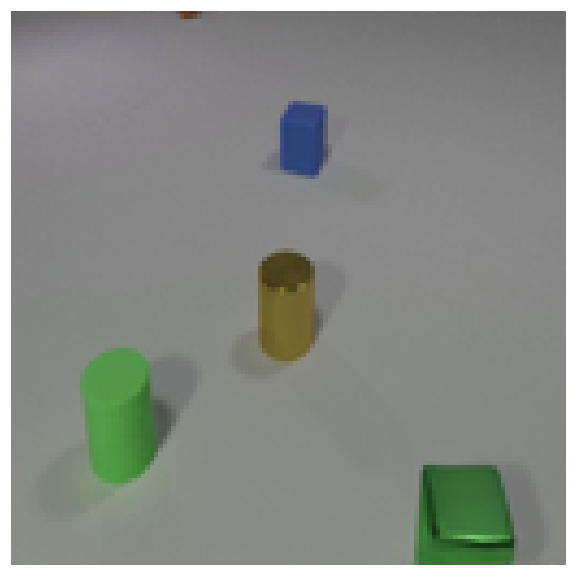} &
        \includegraphics[width=2.5cm]{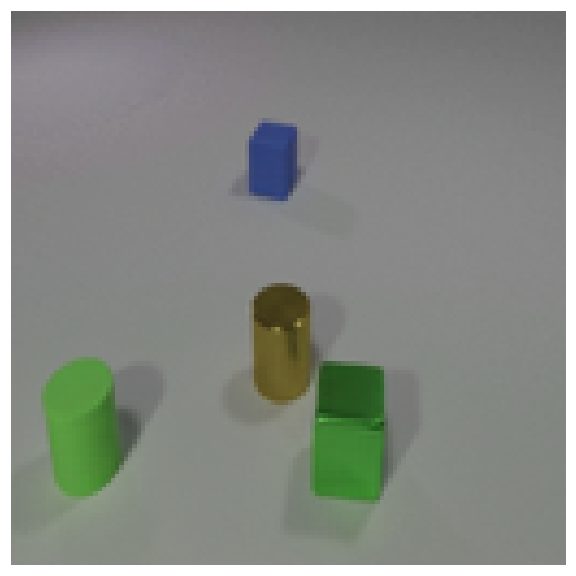} &
        \includegraphics[width=2.5cm]{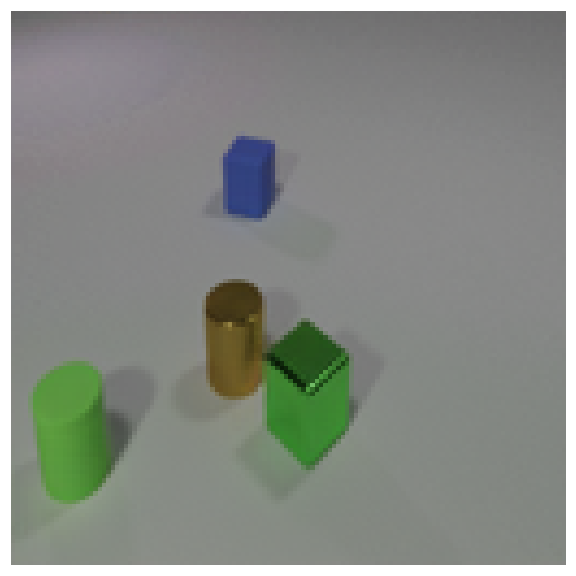} &
        \includegraphics[width=2.5cm]{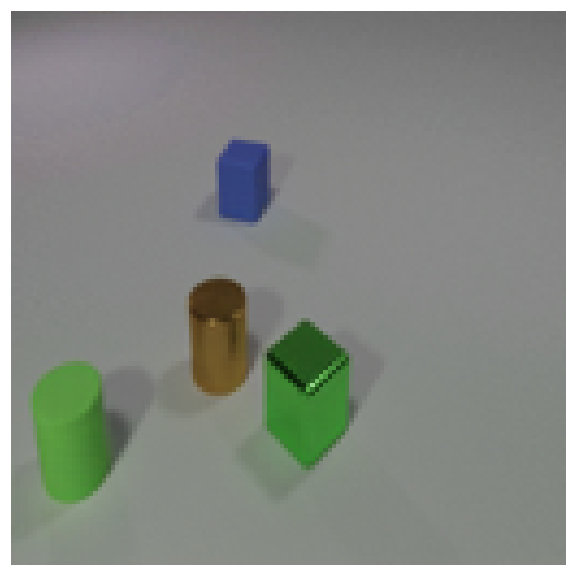} \\
        \animategraphics[width=2.5cm]{7}{Figures/supp/clevrer/5/image_}{0}{60} &  \includegraphics[width=2.5cm]{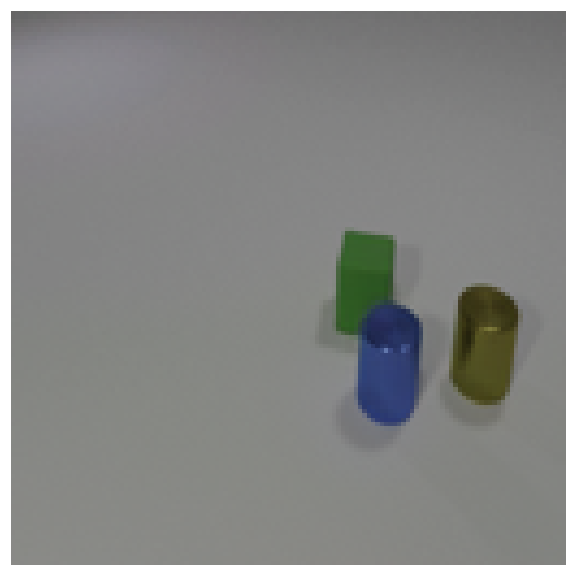} &
        \includegraphics[width=2.5cm]{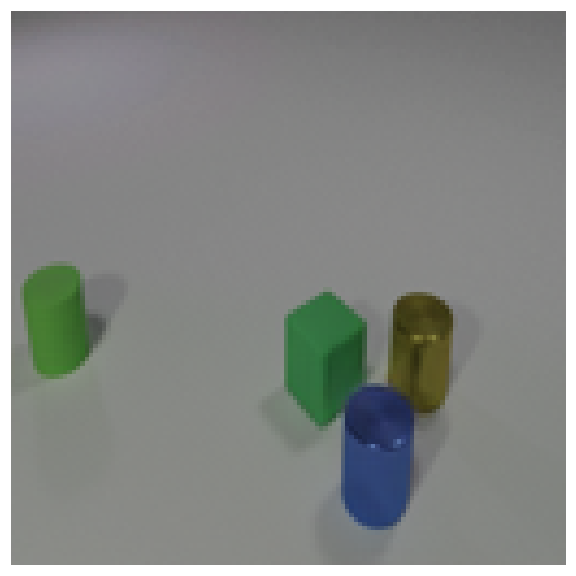} & 
        \includegraphics[width=2.5cm]{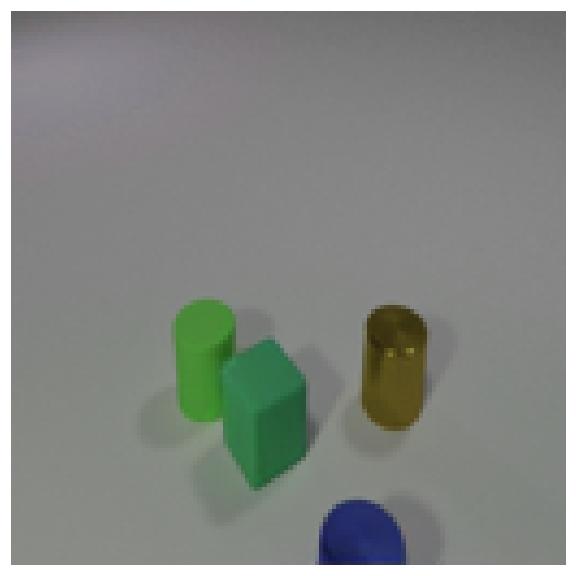} &
        \includegraphics[width=2.5cm]{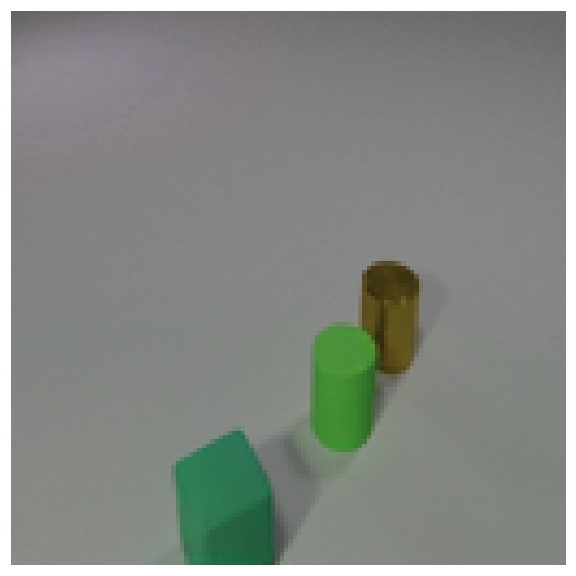} &
        \includegraphics[width=2.5cm]{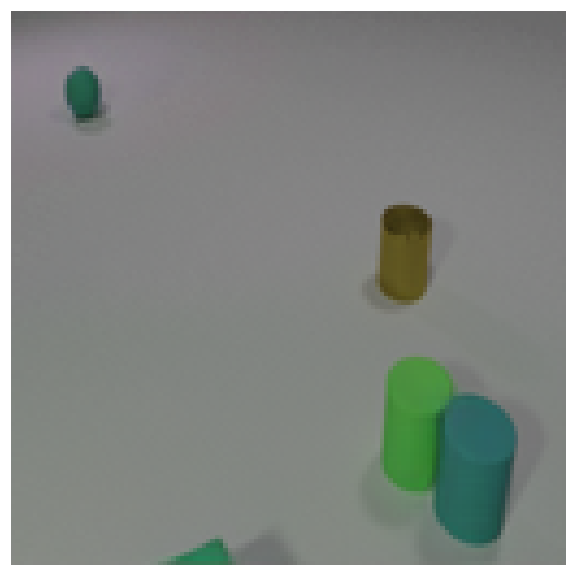} &
        \includegraphics[width=2.5cm]{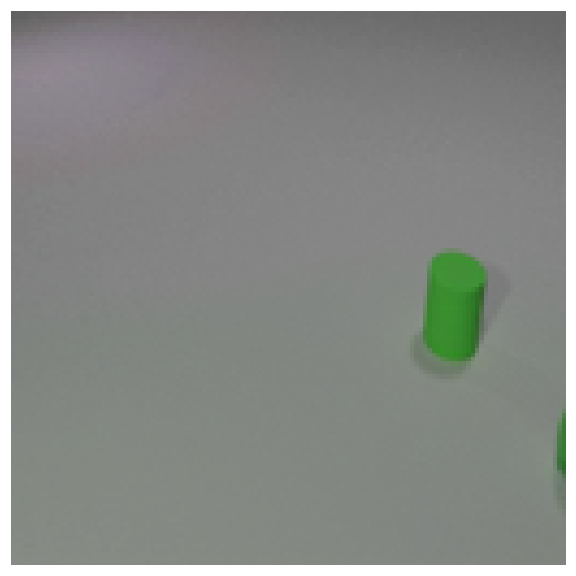} \\
    \end{tabular}
    \caption{Videos generated with \methodName on the CLEVRER~\cite{Yi2020CLEVRERCE} dataset. Please, use Acrobat Reader to play the frames in the leftmost column or open the attached \texttt{index.html} file in your browser.}
    \label{fig:app_clevrer}
\end{figure*}

\begin{figure}[t]
    \centering
    \begin{tabular}{@{}c@{\hspace{0.5mm}}c@{\hspace{0.5mm}}c@{\hspace{0.5mm}}c@{}}
    {\animategraphics[width=2.1cm]{3}{Figures/iper_p1/image_}{0}{5}} &
    {\includegraphics[width=2.1cm]{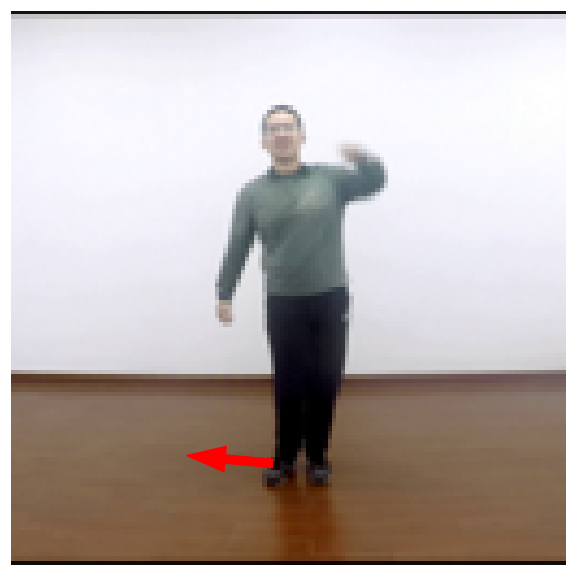}} & 
    {\includegraphics[width=2.1cm]{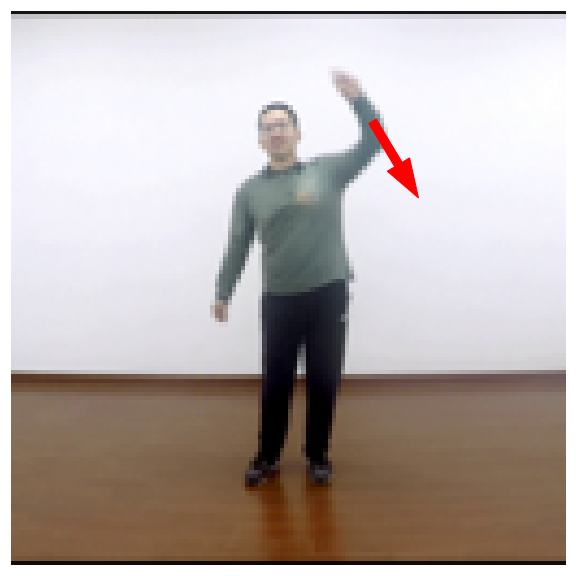}} &
    {\includegraphics[width=2.1cm]{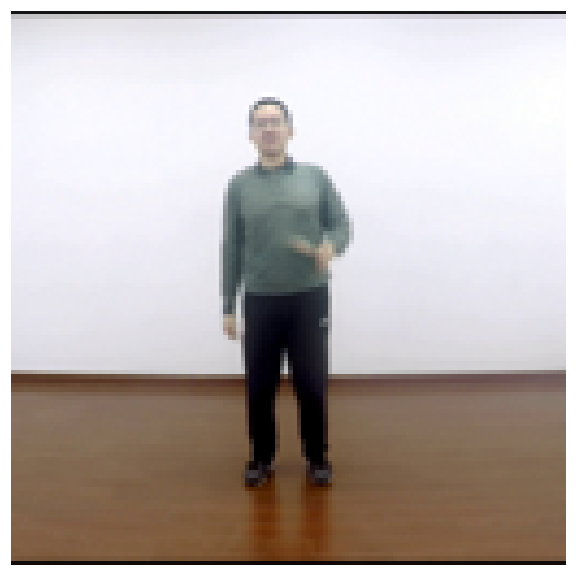}} \\
    {\animategraphics[width=2.1cm]{3}{Figures/iper_p2/image_}{0}{6}} &
    {\includegraphics[width=2.1cm]{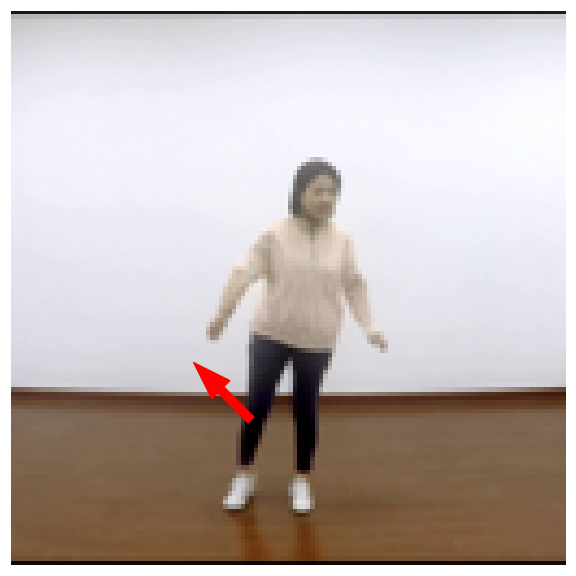}} & 
    {\includegraphics[width=2.1cm]{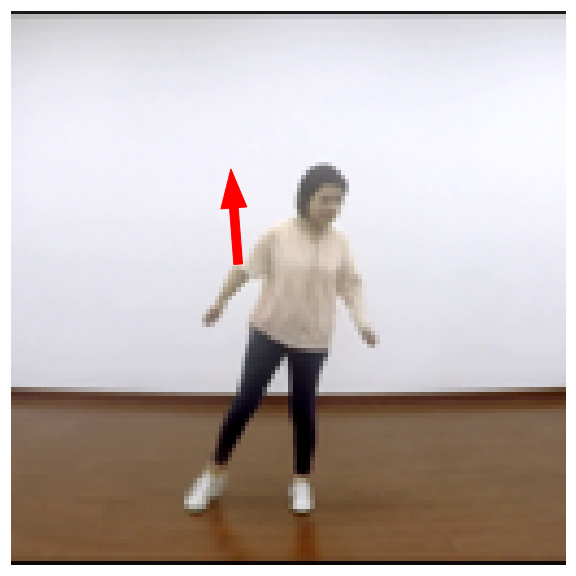}} &
    {\includegraphics[width=2.1cm]{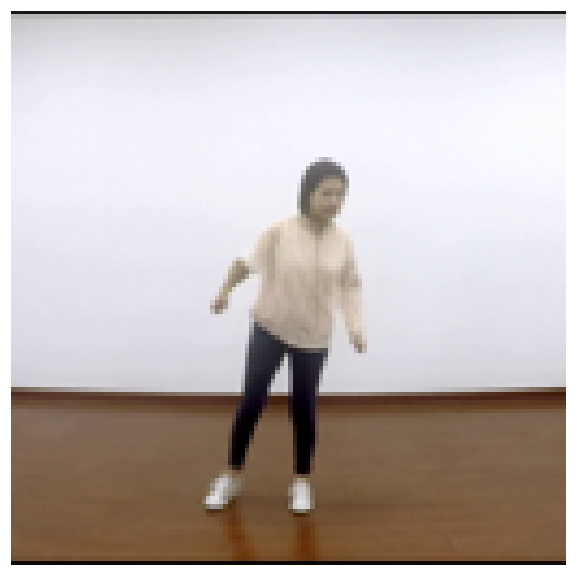}} \\
    \end{tabular}
    \caption{Examples of \methodName simulating human motion on iPER. The images in the first column can be played as videos in Acrobat Reader.}
    \label{fig:iper}
\end{figure}

In this section we provide more qualitative results with \methodName. For videos, please, visit the project's website\footnote{\url{https://araachie.github.io/yoda}}. Figures~\ref{fig:bair} and \ref{fig:app_clevrer} contain some selected sequences to demonstrate the quality of separate frames, which is not visible in the videos on the webpage due to the compression. Figure~\ref{fig:bair} contains more object manipulation scenarios on the BAIR~\cite{Ebert2017SelfSupervisedVP} dataset. Figure~\ref{fig:app_clevrer} provides generated videos capturing long-range consequences of the input controls on the CLEVRER~\cite{Yi2020CLEVRERCE} dataset. Notice the ability of \methodName to realistically model the physics of the scene. \methodName was designed to make it possible to intervene into the generation process at any timestamp, which allows communicating new impulses to the objects on the fly. Figure~\ref{fig:iper} contains results on the iPER~\cite{liu2019liquid} dataset, demonstrating the ability of \methodName to model human motion.

Figure~\ref{fig:arch} provides videos of ablation of the sparse optical flow encoder.

\begin{figure}[t]
    \centering
    \begin{tabular}{@{}c@{\hspace{0.3mm}}c@{}}
    {\animategraphics[width=4cm]{3}{Figures/rebuttal/bair_global/image_}{0}{6}} &
    {\animategraphics[width=4cm]{3}{Figures/rebuttal/bair_ours/image_}{0}{6}} \\
    Convolutional & Ours \\
    \end{tabular}
    \caption{The effect of our encoder with restricted receptive field. Notice, how the motion of the robotic hand is correlated to the input control in the case of convolutional encoder. To play videos, use Acrobat Reader.
    }
    \label{fig:arch}
\end{figure}

\begin{figure}[t]
    \centering
    \begin{tabular}{@{}c@{\hspace{0.5mm}}c@{\hspace{0.5mm}}c@{}}
         \includegraphics[width=2.8cm]{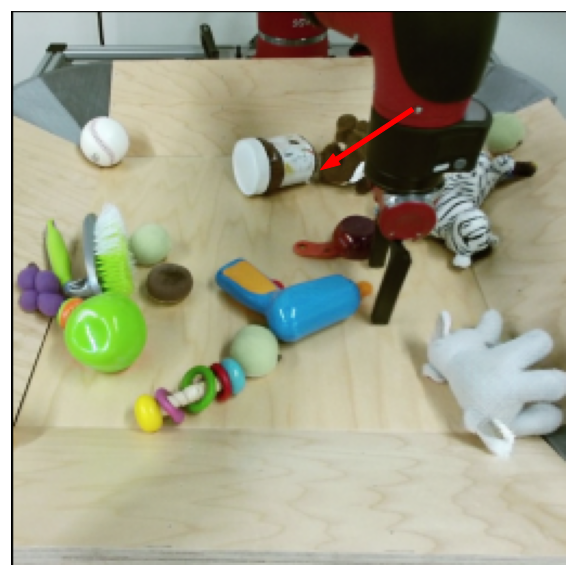} &
         \raisebox{.5mm}{\includegraphics[width=2.8cm,height=2.7cm]{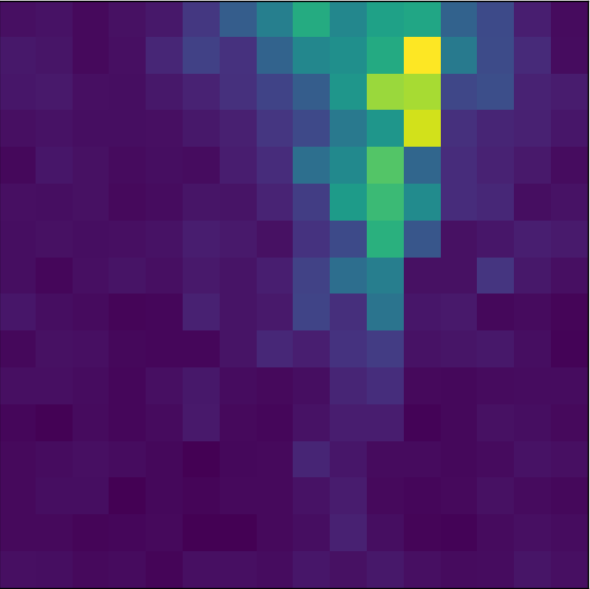}} & 
         \includegraphics[width=2.8cm]{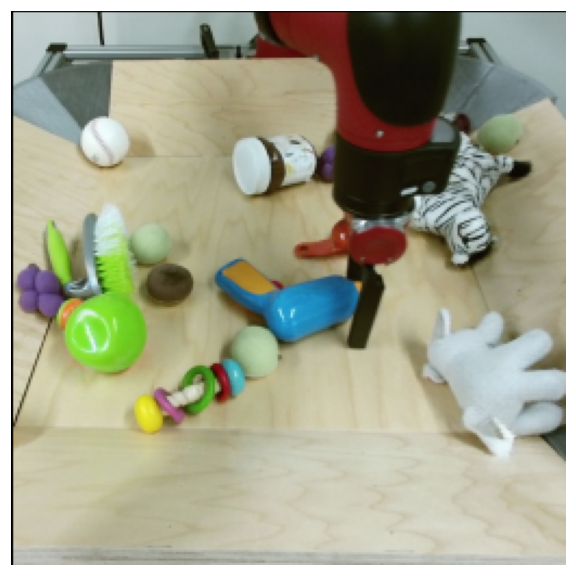} \\
         \includegraphics[width=2.8cm]{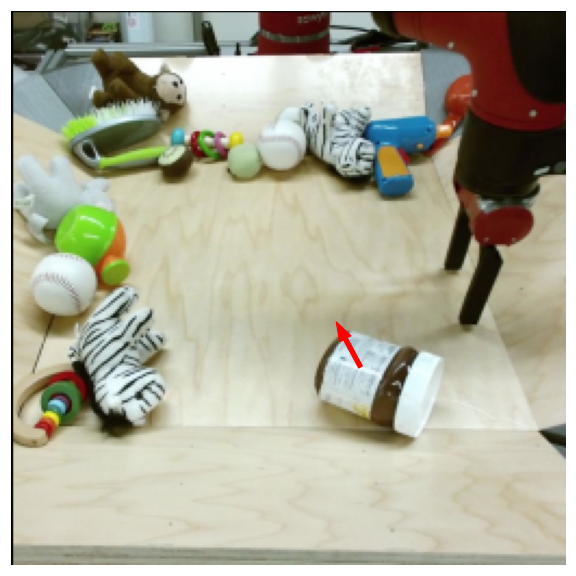} &
         \raisebox{.5mm}{\includegraphics[width=2.8cm,height=2.7cm]{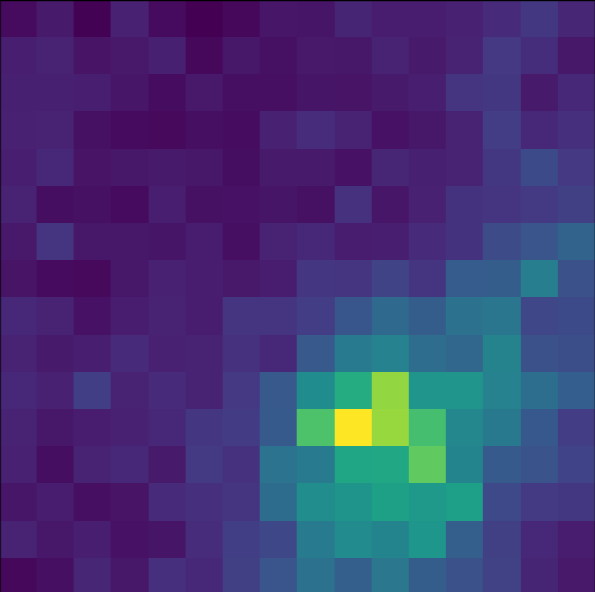}} & 
         \includegraphics[width=2.8cm]{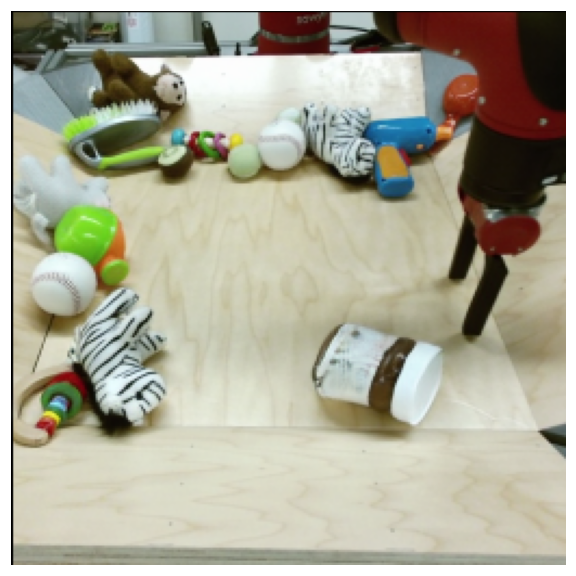} \\  
    \end{tabular}
    \caption{The attention maps from the last layer of the vector field regressor between the interacted location and the rest of the image, averaged over the heads and the flow integration path. It can be seen that the attention maps roughly cover the extents of the objects.}
    \label{fig:bair_attn}
\end{figure}

Since our model is based on Transformers~\cite{NIPS2017_3f5ee243}, we report the attention maps from the last layer of the network between the interacted location and the rest of the image (see middle column in Figure~\ref{fig:bair_attn}). These attention maps often correspond to coarse segmentations of the controlled objects.

\section{Demo}\label{sec:demo}

To ease the access to our results, we designed a demo of \methodName. The user can load the pretrained models and directly interact with the scene by dragging objects with the mouse. The model responds by generating a subsequent frame taking into account the controls specified by the user as well as the previously generated images (see the project's website). The demo and the code are available at the project's github repository\footnote{\url{https://github.com/araachie/yoda}}.

\section{Applications}\label{sec:app}

Pretrained \methodName generates realistic responses to control. This opens up the opportunity to interact with the scene in a counterfactual way, which makes it possible to apply \methodName to downstream tasks. In theory, by optimizing the control inputs to match the given video, one can solve planning or compress the video to a single frame and a sequence of controls. Here we discuss another possible application, object segmentation.

\begin{figure*}[t]
    \centering
    \begin{tabular}{c}
        \includegraphics[width=0.8\linewidth]{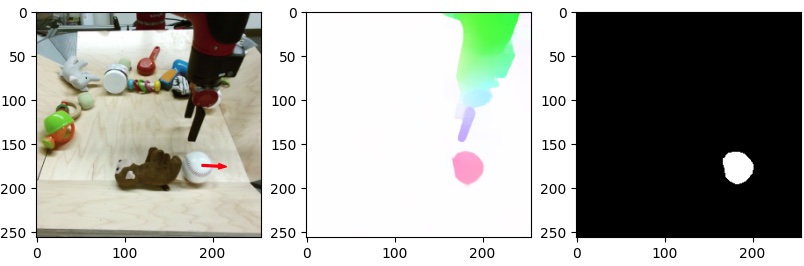} \\
        \includegraphics[width=0.8\linewidth]{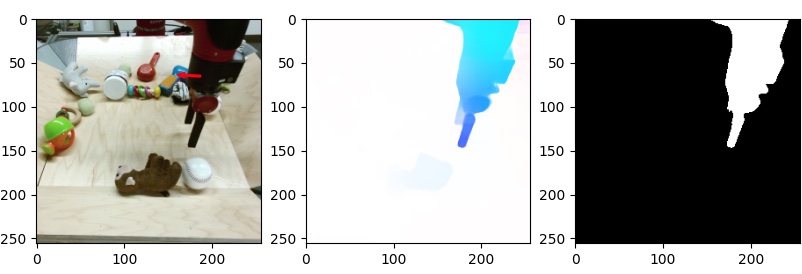}
    \end{tabular}
    \caption{From left to right: the input image, the optical flow between the input image and the next one, the estimated mask. The red arrows in the images in the first column show the control input.}
    \label{fig:segm}
\end{figure*}

Given an image $x$ of a scene with multiple objects and \methodName trained on videos capturing that scene, the user selects a pixel on the object of interest. A random control is then applied to that location to generate the response $x'$ with \methodName. The optical flow field $w$ between $x$ and $x'$ is calculated. All the vectors in $w$ are compared with the input control using some similarity measure (can be $l_2$ or cosine distance). We then apply a threshold to the result to obtain the segmentation mask. For robustness, this procedure can be repeated multiple times and the union of the resulting masks can be used as the final estimator of the segmentation mask. For an example, see Figure~\ref{fig:segm}.

\bibliography{references}

\end{document}